\documentclass[UTF8]{article}
\usepackage[accepted]{icml2025}

    \PassOptionsToPackage{numbers, compress}{natbib}

\usepackage[utf8]{inputenc}

\usepackage{amsmath,amsthm, amssymb}
\usepackage{latexsym}
\usepackage{graphicx}
\usepackage{ifthen}
\usepackage{subcaption}
\usepackage{multicol}
\usepackage{algorithm, algorithmic}
\usepackage[numbers]{natbib}
\usepackage{mathtools}
\usepackage{dsfont}
\usepackage{nicefrac}
\usepackage{subcaption}
\usepackage{enumitem}
\usepackage{bbold}
\usepackage{bm}
\usepackage{booktabs}
\usepackage{CJKutf8}% for chinese characters

\usepackage{bbding}
\usepackage{pifont}
\usepackage{wasysym}
\usepackage{amssymb}

\usepackage{siunitx}

\usepackage[normalem]{ulem}
\useunder{\uline}{\ul}{}
\usepackage{lscape}
\usepackage{tabularx}
\usepackage{wrapfig}

\usepackage{stfloats}  % helps place figures that span both columns exactly where we want, otherwise [b] fails.
\usepackage{float}       % cam ready for 3 algos
\usepackage{placeins}    % cam ready for 3 algos

\FloatBarrier  % prevent earlier floats from jumping in front

\usepackage{hyperref}
\hypersetup{colorlinks=true,linkcolor=blue,citecolor=blue,urlcolor=blue}

\newtheorem*{theorem*}{Theorem}
\newtheorem*{definition*}{Definition}

\newtheorem{theorem}{Theorem}

\usepackage{thm-restate}

\renewcommand{\algorithmiccomment}[1]{\bgroup\hfill\footnotesize~#1\egroup}

\newcommand{\sample}{\textsc{Sample}}
\newcommand{\sequence}{\textsc{Sequence}}
\newcommand{\geom}{\textsc{Geom}}

\DeclareMathOperator{\argmax}{argmax}

\newcommand{\OPT}{\texttt{OPT}}

\newcommand{\D}{\mathcal{D}}

\icmltitlerunning{E-LDA: Toward Interpretable LDA Topic Models with Strong Guarantees in Logarithmic Parallel Time}

\begin{document}

\twocolumn[
% \icmltitle{Submission and Formatting Instructions for \\
%            International Conference on Machine Learning (ICML 2025)}
% \icmltitle{Toward Interpretable LDA Topic Models with Strong Guarantees\\
%             in Logarithmic Parallel Time}

\icmltitle{E-LDA: Toward Interpretable LDA Topic Models with Strong Guarantees\\
            in Logarithmic Parallel Time}

\icmlsetsymbol{equal}{*}

\begin{icmlauthorlist}
\icmlauthor{Adam Breuer}{yyy}
\end{icmlauthorlist}
\icmlaffiliation{yyy}{Department of Computer Science, Dartmouth College, Hanover, NH, USA}
\icmlcorrespondingauthor{Adam Breuer}{adam.breuer@dartmouth.edu}

\icmlkeywords{Topic Model, LDA, Interpretability, Social Science, Causal Inference}

\vskip 0.3in
]



\printAffiliationsAndNotice{}  % leave blank if no need to mention equal contribution
\begin{abstract}
In this paper, we provide the first practical algorithms with provable guarantees for the problem of inferring the topics assigned to each document in an LDA topic model. This is the primary inference problem for many applications of topic models in social science, data exploration, and causal inference settings. We obtain this result by showing a novel non-gradient-based, combinatorial approach to estimating topic models. This yields algorithms that converge to near-optimal posterior probability in logarithmic parallel computation time (adaptivity)---exponentially faster than any known LDA algorithm.  We also show that our approach can provide interpretability guarantees such that each learned topic is formally associated with a known keyword. Finally, we show that unlike alternatives, our approach can maintain the independence assumptions necessary to use the learned topic model for downstream causal inference methods that allow researchers to study topics as treatments. In terms of practical performance, our approach consistently returns solutions of higher semantic quality than solutions from state-of-the-art LDA algorithms, neural topic models, and LLM-based topic models across a diverse range of text datasets and evaluation parameters.
\end{abstract}

\section{Introduction}
\label{sec:intro}
%
% %
%
%
%
%
Topic models such as Latent Dirichlet Allocation (LDA) \citep{blei2003latent} have become central to numerous high-profile streams of social science research ranging from the history of science to economics and political science \citep{ash2023text, barbera2019leads, 
bybee2024business, 
djourelova2023persuasion, 
i2012revolving,
 catalinac2016pork,
griffiths2004finding,
hall2008studying, 
lauderdale2014scaling,
martin2019local, 
dietrich2019pitch,
mueller2018reading, 
pan2018concealing, 
tingley2017rising}. 
%larsen2019value,
% \citep{butler2023have, wratil2022government, rossiter2022measuring, kennard2022controls, 
% motolinia2021electoral, yoder2020does, martin2019local, barbera2019leads, mueller2018reading, pan2018concealing, barnes2018making, brutger2018dispositional, kim2017political, tingley2017rising, catalinac2016pork, huff2016banners, lauderdale2014scaling, i2012revolving, wang2009human, hall2008studying, fei2005bayesian, griffiths2004finding}
These models seek to learn the latent thematic structure that describes the content of a  text dataset. Specifically, topic models posit that each document's words were drawn from a (latent) document-specific mixture of (latent) \emph{topics}, where each topic is a distribution over all words in a vocabulary. 

% CAM READY kim2017political,

For many applications of topic models, the primary goal is to estimate the set of latent topics, which provides a simple yet semantically rich summary of the data. In contrast,  for the majority of applications in social science and data exploration, practitioners' primary goal is to infer each document's most probable assigned topics. These document-level topic assignments are then used as document labels that can be piped into downstream document-level regression analysis or causal inference methods.

Social scientists continue to strongly prefer LDA for this task over recent neural and LLM/BERT-based topic models for two reasons. First, social science research designs favor parametric models that are amenable to analysis. This is particularly important when using topic model outputs as labels for downstream models whose theoretic guarantees require that labels are well-estimated, consistent, conditionally independent, etc. Second, social scientists assume that the latent LDA topic space is semantically interpretable. However, there are three key challenges. 

\textbf{There are no known provable guarantees for the topic assignment problem.} First, it is well-known that full posterior inference of LDA parameters is NP-hard even with just $2$ topics \citep{arora2012learning}. Under various assumptions, it is possible to approximate LDA's set of topics in poly-time \citep{anandkumar2012spectral, arora2012learning, arora2013practical, arora2016computing, arora2018learning}. However, even if these latent topics are known, it is still NP-hard in general to learn LDA's Maximum A Posteriori (MAP) assignments (i.e., which topic probably generated each word in a given document) \citep{sontag2011complexity}. Practitioners therefore resort to heuristic inference techniques such as variational inference \citep{blei2003latent} or Gibbs samplers \citep{griffiths2004finding}. These approaches are well-motivated, but they seek only locally modal solutions and produce poorly fit results on some datasets \citep{gerlach2018network}. %

% \textbf{Existing algorithms are generally incompatible with causal inference.} Second, there has been much recent interest in causal inference frameworks that consider documents' topics as treatments or outcomes \citep{egami2022make, feder2021causalm, feder2022causal, fong2016discovery, fong2023causal, hu2021causal, veitch2020adapting}. However, obtaining documents' topic assignments for these frameworks requires a 2nd round of out-of-sample inference algorithms that are incompatible with the usual guarantees of causal inference, and that are known to yield unbounded errors, bias, and spurious causal inferences \citep{battaglia2024inference}.
% %pryzant2018deconfounded, wood2018challenges % PUT IN CAM READY

\textbf{Existing algorithms are generally incompatible with causal inference.} Second, there has been much recent interest in causal inference frameworks that consider documents' topics as treatments or outcomes \citep{egami2022make, feder2021causalm, feder2022causal, fong2016discovery, fong2023causal, hu2021causal, veitch2020adapting}. Obtaining documents' topic assignments for these frameworks requires a 2nd round of inference using out-of-sample inference algorithms. However, out-of-sample algorithms also lack approximation guarantees, which undermines the usual guarantees for causal inference and yields spurious inferences with unbounded bias \citep{battaglia2024inference}.
%pryzant2018deconfounded, wood2018challenges % PUT IN CAM READY

\textbf{Existing algorithms learn and assign some uninterpretable topics.}  Finally, social scientists continue to prefer parametric topic models like LDA over neural and LLM-based models because they assume that the latent LDA topic space is interpretable. However, practitioners who wish to understand LDA results must manually parse topics that are arbitrary distributions over e.g. $20{,}000$ words in search of semantic meaning, then subjectively choose a topic `label' (e.g. ``this topic is about `physics'")---a burdensome and subjective process that was famously compared to `reading tea leaves' \citep{chang2009reading, lau2014machine}. 
%This is because there is no formal relationship between a topic `label' and the probability mass that the topic assigns to any of the e.g. $20{,}000$ words in the vocabulary\footnote{Topic model inference algorithms are also highly complex, so a human cannot follow their steps to vet \emph{why} the algorithm generated a given topic or \emph{why} it associated a particular topic with a given document.}. A standard social science research design depends on numerous such labelling choices, which is fraught when each topic may be mis-approximated, when the assignments of topics to documents are arbitrarily poorly fit in the absence of provable guarantees, and when available algorithms are too complex for researchers to vet why e.g. a given topic was assigned to a document. 
%
In practice, researchers find that 10\% of the topics learned by standard algorithms are fully nonsensical, and many more are partially nonsensical \citep{mimno2011optimizing}. These problems have led to vast research on post-hoc evaluation of topics' semantic quality and inspired recent excitement about algorithms that attempt to heuristically associate topics with keywords \citep{eshima2020keyword, doshi2015graph,  mimno2011optimizing,  roder2015exploring}. However, \emph{we lack a formal means to guarantee that each topic ascribes probability mass in an interpretable way, such as via a known function of a known keyword.}
    % newman2010automatic for cam ready

% \textbf{Existing algorithms violate assumptions required for causal inference.} Very recently, there has also  been a great deal of interest in causal inference frameworks that consider texts' topics as treatments or outcomes \citep{egami2022make, feder2021causalm, feder2022causal, fong2016discovery, fong2023causal, hu2021causal, pryzant2018deconfounded, veitch2020adapting, wood2018challenges}. The LDA model itself is compatible with the independence assumptions (SUTVA, etc.) required for these downstream causal inference frameworks. However, all known LDA algorithms violate these assumptions because the topics they assign to one document depend on all other documents. Researchers have recently developed various new techniques to overcome this problem, but they require multiple rounds of complex inference algorithms that are incompatible with the usual provable guarantees of causal inference methods, and they are known to introduce additional and unbounded approximation errors \citep{egami2022make}.

\textbf{These various problems make LDA an outlier among related ML techniques.} Seminal work of \citet{blei2003latent} shows that LDA is a hierarchical extension of clustering (e.g. `mixture of unigrams').
%\footnote{Namely, in clustering i.e. `mixture of unigrams' models, all words in one document are drawn from the same topic, so a collection of documents is thus a mixture. Topic models such as LDA simply add a latent step to the hierarchy, so each \emph{document} is a (latent) document-specific mixture of unigrams (topics).} 
Yet unlike LDA, clustering models have recently benefited from theoretic breakthroughs in combinatorial optimization that yielded fast algorithms with near-optimal approximation guarantees. These include exemplar-based problem formulations that convey additional interpretability advantages and other desirable properties \citep{breuer2020fast, balkanski2019optimal, liu2013entropy, mirzasoleiman2013distributed, mirzasoleiman2015lazier}.
%lashkari2007convex add to cam ready
\vspace{0.1mm}
\begingroup
\addtolength\leftmargini{-0.1in} % shrink quote margin
\begin{quote}
\vspace{-1.4mm}
\vspace{-1.5mm}
\noindent\hfill$\diamond\ \diamond\ \diamond$\hfill\null\\
% \noindent\hfill$\blacklozenge\ \blacklozenge\ \blacklozenge$\hfill\null\\
% \noindent\hfill$\bullet\ \bullet\ \bullet$\hfill\null\\
% \noindent\hfill\textbf{$\diamondsuit\ \diamondsuit\ \diamondsuit$}\hfill\null\\
%\vspace{-1mm}
\emph{Can the breakthroughs in optimization that recently revolutionized clustering \mbox{algorithms} also yield similarly fast, near-optimal, interpretable topic models?}
\vspace{-1mm}
\noindent\makebox[\linewidth]{\hfill$\diamond\ \diamond\ \diamond$\hfill}
\vspace{-5mm}
 \end{quote}
 \endgroup
% \begin{center}
% \huge{$\dots$}\normalsize
% \end{center}
%\vspace{-5mm}

% \vspace{-1em}
% \begingroup
% %\setlength\leftmargini{0em} % remove any extra quote margin
% \addtolength\leftmargini{-0.1in} % shrink quote margin
% \noindent\hfill\rule{2cm}{0.4pt}\hfill\null
% \vspace{-1.5mm}
% \begin{quote}
% \vspace{-1.5mm}
% \emph{Can the breakthroughs in optimization that recently revolutionized clustering \mbox{algorithms} also yield similarly fast, near-optimal, interpretable topic models?}
% \end{quote}
% \vspace{-1em}
% \noindent\hfill\rule{2cm}{0.4pt}\hfill\null
% \vspace{-2em}
% \endgroup

\vspace{-1em}
\textbf{Our contribution.}
We show the first practical algorithms that near-optimally solve the Maximum A Posteriori (MAP) topic-word assignment problem for LDA---that is, we provably obtain the near-optimal topic from which each word (and document) in the dataset was probably drawn. 

We contrast our approach with standard LDA algorithms, which fix a small set of topics and use gradients to adjust them incrementally towards a local mode. That standard approach relies on hyperparameters to encourage a desirably sparse set of topics in document's underlying distributions. Then, after fitting a model, practitioners typically compute topics' coherence scores, prune low-coherence topics, and manually label topics with heuristic keywords.

We invert this approach to develop a combinatorial approach to estimating topic models. Suppose we have a very large ‘candidate set’ of topics from which the sparse set of topics in the solution are selected. It is easy to obtain this candidate set from other algorithms. Alternatively, we show we can simply initialize it to \emph{all} feasible topics that meet standard interpretability criteria, so any topic selected for the solution will ascribe mass via a known function of some keyword. Then, instead of relying on gradients and heuristic sparsity-encouraging hyperparameters, we proceed combinatorially. Specifically, we consider the MAP assignment problem under an explicit sparsity constraint that upper-bounds the mean number of topics from which each document was drawn. This approach conveys multiple theoretic advantages for social science applications as we discuss below, and it brings LDA into line with recent techniques in clustering and data summarization. 

\noindent\textbf{Provable guarantees}. Our algorithm obtains a \emph{deterministic} worst-case $1$-$1/e$ approximation of the constrained log MAP probability objective (i.e. the most probable topic from which each word was drawn) in a number of iterations that is linear in the count of documents, where each iteration's runtime is just \emph{logarithmic} in the count of documents. We also give a parallel algorithm that converges w.p. in \emph{logarithmic} iterations (adaptivity). This is exponentially faster than any known LDA algorithm. Because our algorithms can also perform out-of-sample inference, they also provide a means to use LDA for causal inference with provable approximation guarantees.

%There are three main applications. First, given topics inferred by other algorithms (including those with guarantees), our algorithms obtain provably good MAP assignments. Second, because our technique allows us to define the `candidate set' of topics from which assignments are selected, we can optionally guarantee that our solution is interpretable by initializing this set to all possible topics that meet standard interpretability criteria. The result is a new, non-gradient-based way to learn interpretable topic models that is intuitively familiar to the human process of connecting documents to the topics that best describe them. 

% . For example, it is easy to initialize this set so that each candidate topic assigns mass to each word in the vocabulary in proportion to its co-occurrence with some keyword, and the candidate set includes a topic for each possible keyword in the vocabulary. In addition, because our algorithms can enforce SUTVA and other properties, they also provide the first means to use LDA for downstream causal inference with provable approximation guarantees.

\noindent\textbf{Practical performance}. Finally, we show that our algorithms consistently learn solutions of higher semantic quality (coherence) than those found by state-of-the-art LDA algorithms, neural topic models, and LLM/BERT-based topic models. These improvements hold across a diverse set of text datasets and evaluation parameters.

% Our algorithms also obtain better topic-word assignments in terms of posterior
% probability even when constrained to use the Gibbs topics
% and keep the same avg. sparsity as the Gibbs assignments.

\textbf{Technical overview.} From a purely technical perspective, there are four major challenges addressed in this work. The first is to show that under certain conditions, the classic LDA MAP assignment problem can be cast as a combinatorial optimization problem whose objective function exhibits the key diminishing returns property known as monotone submodularity. Submodularity is heavily studied in clustering, data summarization, and NLP problems, but surprisingly, its connection to topic models remained undiscovered \citep{gomes2010budgeted, mirzasoleiman2015lazier, mirzasoleiman2016fast, lin2011class}. Due to this property, the problem admits near-optimal solutions via generic Greedy algorithms \citep{NWF78}. Unfortunately, vanilla Greedy algorithms are practically infeasible for this objective because their runtime complexity scales as a third order polynomial in terms of the number of documents. In fact, because this objective is defined on sets of topic-to-document connections (links), not documents, solving it for even modest text corpora requires us to solve some of the largest combinatorial optimization problems considered in any academic research in terms of the sizes of the ground set (potential links) and solution set. Therefore, our second challenge is to design a fast serial algorithm for this problem with the same near-optimal Greedy approximation guarantee by leveraging the special independence structure of LDA's probabilistic model. We then show that our techniques also apply to very recent parallel adaptive sampling algorithms \citep{balkanski2018adaptive, breuer2020fast}. We leverage this to obtain near-optimal MAP solutions w.h.p. in logarithmic parallel time.

Because our techniques are combinatorial, they permit the researcher to define the (possibly very large) `candidate set' of topics from which the solution topics will be selected. This set can be obtained from existing algorithms, but this does not address the interpretability problems described above. Therefore, our third challenge is to show that it is possible to \emph{efficiently generate} this candidate set of topics via standard coherence functions that are widely used to \emph{measure} topics' interpretive quality ex-post. Finally, our fourth challenge is to show that our techniques enforce document independences (SUTVA, etc.) required for downstream causal inference frameworks. This enables causal inference research designs on documents' LDA topics that inherit near-optimal approximation guarantees.

\vspace{-0.2em}

% \section{The LDA model and the MAP topic-word assignment problem}
\section{LDA MAP topic-word assignment problem}
\label{sec:LDAbackground}
The LDA model states that the data is a corpus (set) $D$ of documents and each document $d\in D$ is a multiset of words: $d=[w_1, \dots, w_{|d|}]$. Suppose there are $k$ topics, where each topic is a distribution over all words in the vocabulary. LDA states that the text corpus is generated as follows: For each document $d$, we sample a document length $|d|$ and a document mixture distribution $\theta_d\sim Dirichlet (\alpha)$ over the $k$ topics. Then, to draw each of the $|d|$ words in $d$, we sample a topic $z_{d,i}\in \{1,\dots,k\}$, where $z_{d,i}\sim Categorical(\theta_d)$, and from this topic we sample a word $w_{d,i}\sim Categorical(\phi_{z_{d,i}})$, where $\phi_{z}$ is the distribution over all words corresponding to topic $z$. 
%
%
%
%
%
% \begin{figure*}[b]  % THE * DISPLAYS IT ACROSS BOTH COLUMNS
% \label{illustration}
% %\centering
% \begin{subfigure}{1.0\textwidth}
% \centering
% \includegraphics[width=0.67\textwidth]{opttopic_diagrams/GRAPHICS_FOR_OPTTOPIC_DISSERTATION_iteration5v2 fixed iter5.pdf} % ITER1 ITER5 NAMES ARE FLIPPED 
% %
% \caption[E-LDA after the \textbf{1st} Greedy iteration.]{\emph{Above:} Function $\dot{f}$ after the \textbf{1st} Greedy iteration (1 topic-to-document link added to solution set $E$, i.e., the arrow between the China topic and document $1$). Note that the candidate topics $\Phi$ are each represented by their `label word' in the left column (see Section \ref{sec:candidates}).}
%    \label{fig:diagiter1} 
% \end{subfigure}
% \vspace{-2em}  % Reduces vertical spacing between subfigures
% \begin{subfigure}{1.0\textwidth}
% \centering
% \includegraphics[width=0.67\textwidth]{opttopic_diagrams/GRAPHICS_FOR_OPTTOPIC_DISSERTATION_iteration5v2 fixed iter1.pdf} % ITER1 ITER5 NAMES ARE FLIPPED 
% \caption[E-LDA after the \textbf{5th} iteration]{$\dot{f}$ after \textbf{5th} iteration. Each word must be drawn from some topic connected to its document (note colors).}
%    \label{fig:diagiter5}
% \end{subfigure}
% \vspace{-1em} % if those 2 lines dont squeeze in here then they bump like 6 lines later on.
% %\caption{Diagram of }
% %\label{fig:mydiagramsICML}
% \end{figure*}
%
%
%
\begin{figure*}[b]  % THE * DISPLAYS IT ACROSS BOTH COLUMNS
\vspace{-1em} 
\label{illustration}
%\centering
\begin{subfigure}{1.0\textwidth}
\centering
\includegraphics[width=0.65\textwidth, height=0.24\textwidth]{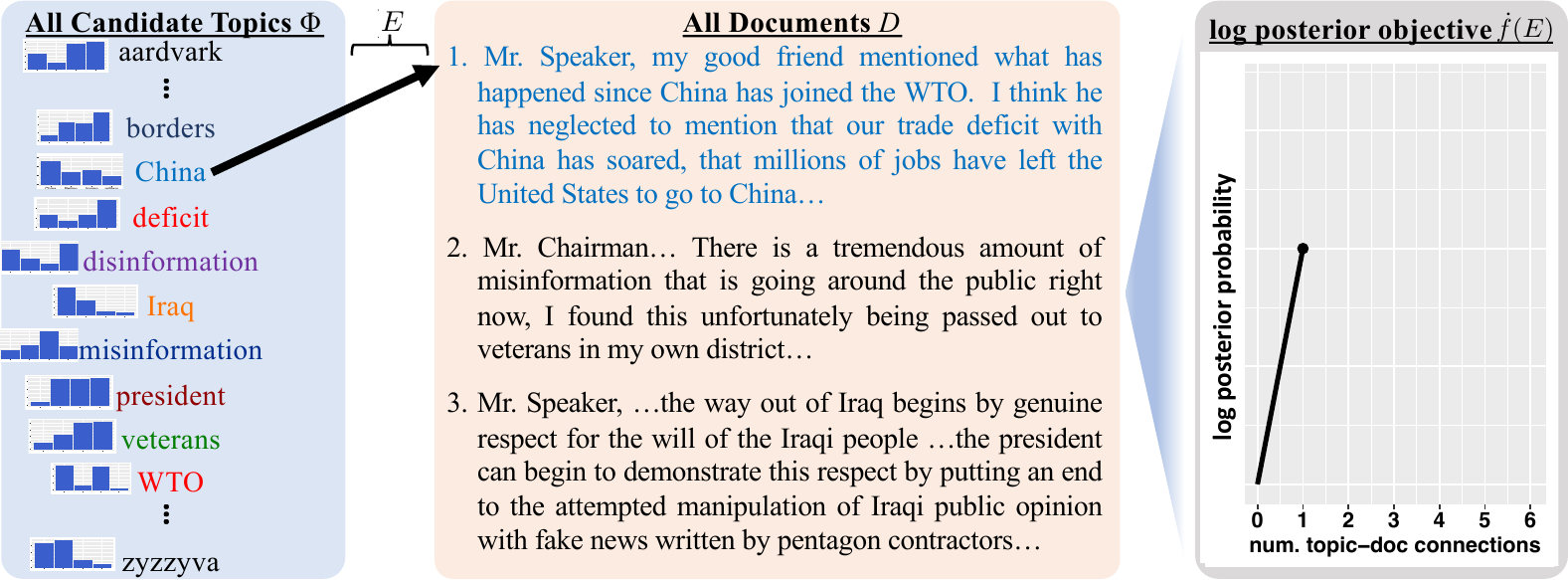} % 
%
% \caption[E-LDA after the \textbf{1st} Greedy iteration.]{\emph{Above:} Function $\dot{f}$ after the \textbf{1st} Greedy iteration (1 topic-to-document link added to solution set $E$, i.e., the arrow between the China topic and document $1$). Note that the candidate topics $\Phi$ are each represented by their `label word' in the left column (see Section \ref{sec:candidates}).}
\caption[E-LDA after \textbf{1} Greedy iteration.]{\emph{Top:} Objective $\dot{f}$ after \textbf{1} Greedy iteration. Note that $1$ topic-to-document link was added to solution set $E$.}%, i.e., the arrow between the China topic and document $1$). Note that the candidate topics $\Phi$ are each represented by their `label word' in the left column (see Section \ref{sec:candidates}).}

   \label{fig:diagiter1} 
\end{subfigure}
%\vspace{-2em}  % Reduces vertical spacing between subfigures
\vspace{-\baselineskip}
\begin{subfigure}{1.0\textwidth}
\centering
\includegraphics[width=0.65\textwidth, height=0.24\textwidth]{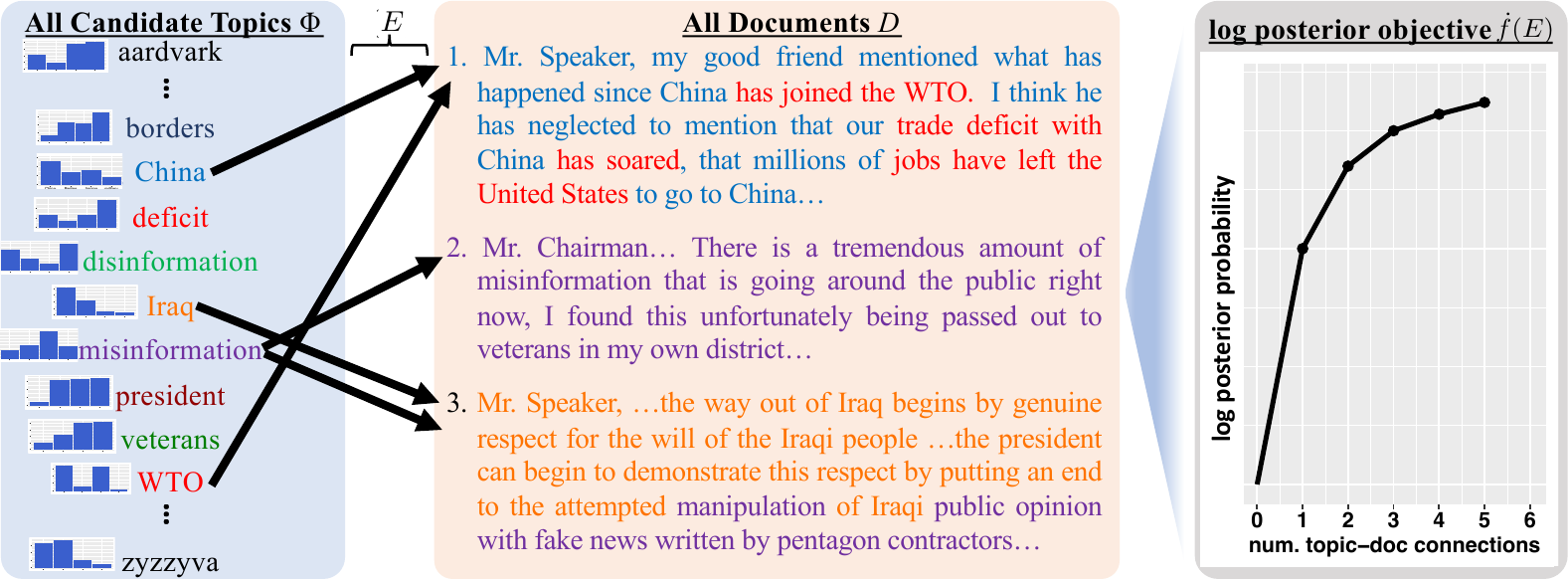} % 
% \caption[\vspace{-0.3em} $\dot{f}$ after the \textbf{5th} iteration]{\emph{Lower figure:} $\dot{f}$ after \textbf{5th} iteration. Each word must be drawn from some topic connected to its document, per eqn. \ref{eq:easy} (note colors).}
\caption[\vspace{-0.3em} $\dot{f}$ after the \textbf{5th} iteration]{\emph{Bottom:} $\dot{f}$ after \textbf{5} iterations. Each word is (re-)assigned (see colors) to its most likely topic of those linked to its document per eq. \ref{eq:easy}.}

   \label{fig:diagiter5}
\end{subfigure}
% \vspace{-1.8em} % if those 2 lines dont squeeze in here then they bump like 6 lines later on.
\caption{\vspace{0.3em} Illustration of \textsc{E-LDA} approach to topic modeling. Note submodular `diminishing returns' in $\dot{f}$.\vspace{-1.96em}}
%\label{fig:mydiagramsICML}
\end{figure*}
We refer to the $k$ topics as \emph{candidate topics} to distinguish them from \emph{realized topics} that are actually sampled. Denote by $\Phi$ the set of candidate topic distributions $\{\phi_1,\dots,\phi_k\}$; by $Z$ all latent realized topics $z_{d,i}$ assigned to each word in $d\in D$; by $\Theta$ all documents' topic mixtures $\{\theta_1,\dots, \theta_{|D|}\}$; and by  $n_{d,\tau}$ the count of words in $d$ that are drawn from topic $\tau$. After integrating $\Theta$ out of the joint model probability, and assuming $\Phi$ is fixed (i.e. topics are already estimated), we obtain the canonical MAP topic-word assignment problem studied by \citet{sontag2011complexity} (App.  \ref{appendix:integratingouttheta}): 
% %
% \begin{align}
%  \log \text{P}(Z|D)\propto  \sum_{d \in D} \Big( &\underbrace{\sum^{|d|}_{i=1} \log Cat (w_{d,i} | z_{d,i},  \phi_{z_{d,i}})}_{likelihood} + \underbrace{\sum^k_{\tau=1} \log \Gamma(n_{d,\tau} + \alpha_\tau) }_{sparsity}\notag \\[-4pt]  &\enspace+\underbrace{\log \Gamma(\sum^k_{\tau=1} \alpha_\tau) - \log \Gamma(\sum^k_{\tau=1} \alpha_\tau + |d|) - \sum^k_{\tau=1} \log \Gamma(\alpha_\tau)}_{constant}
%             \Big)
%          \label{eq:logMAP}
% \end{align}
% %
\begin{align}
\vspace{-1.5em}
\label{eq:logMAP}
 \log & \ \text{P}(Z|D)\propto  \\[-0.5em] \enspace & \sum_{d \in D} \Big( \underbrace{\sum^{|d|}_{i=1} \log Cat (w_{d,i} | z_{d,i},  \phi_{z_{d,i}})}_{likelihood} + \underbrace{\sum^k_{\tau=1} \log \Gamma(\text{$n_{d,\tau}$$+$$\alpha_\tau$}) }_{sparsity}\notag \\[-4pt]   &  \enspace+\underbrace{\log \Gamma(\sum^k_{\tau=1} \alpha_\tau) - \log \Gamma(\sum^k_{\tau=1}\text{$\alpha_\tau$$+$$|d|$}) - \sum^k_{\tau=1} \log \Gamma(\alpha_\tau)}_{constant}
            \Big) \notag
\end{align}
\textbf{MAP assignments are desirable for data exploration \& causal inference.} Maximizing \mbox{eq. \ref{eq:logMAP}} for a set $\Phi$ of topics gives the MAP assignment i.e. the topic $z_{d,i}$ that each word in the dataset was probably drawn from. This is concrete, easy to interpret, and well-defined as a causal treatment. 

\textbf{MAP assignment is NP-hard even for 1 document, and we lack a practical approximation \citep{sontag2011complexity}.} Instead, standard data exploration and causal inference applications resort to using documents' underlying topic mixtures $\theta_d$, which are considered more tractable \citep{sontag2011complexity, arora2016provable}. This is disadvantageous, as interpreting $\theta_d$ requires manually inspecting a per-document $k$-dimensional distribution-over-all topic-distributions (incl. many unrealized topics). Also, $\theta_d$ is ill-defined as a causal treatment \citep{fong2016discovery}.

\section{MAP problem with realized topic sparsity}
\label{sec:ETM}
In this work, we take inspiration from recent techniques in clustering and feature selection \citep{askari2020naive, khanna2017scalable, QS19} to show that LDA's MAP assignment problem is tractable in the infinite limit of documents' topic sparsity prior $\alpha$ under explicit sparsity (cardinality) constraints. In other LDA use-cases that seek documents' underlying topic mixtures $\theta_d$, it is standard to choose a uniform prior over $\theta_d$ by setting $\alpha$=$1$. In contrast, $\theta_d$ are nuisance parameters in our use-case, whereas $\alpha\rightarrow\infty$ corresponds to uniform probability over our parameters of interest: the realized MAP topic assignments from which words were drawn. Thus, we do \emph{not} have the user choose a prior $\alpha$ to heuristically push each document's underlying distribution to favor few topics. Instead, we say (by $\alpha \rightarrow \infty$) that all topics were equally likely a priori (each document's \emph{underlying} topic distribution becomes a point centered on the topic simplex), but for each document only a few topics (e.g., an avg. of $4$ per document) were \emph{actually} drawn. This choice also conveys a linguistic advantage, as the conventional reliance on $\alpha$ is known to drive degenerate unimodal solutions---a main source of linguistic criticism \citep{gerlach2018network, wallach2009rethinking}.
% We now give Theorem \ref{theorem:infinitealpha}, which proves that . We defer the proof to Appendix \ref{appendix:InfiniteLimitProof}.
We now give Theorem \ref{theorem:infinitealpha}, which forms the basis of our approach. We defer the proof to App. \ref{appendix:InfiniteLimitProof}. %We will then show below that the resulting equation is tractable such that we can optimize it to obtain provably near-optimal MAP solutions:
% Theorem 2 proves that
% We now give Theorem \ref{theorem:infinitealpha}, which shows that in this limit $\alpha\rightarrow\infty$,  $\log \text{P}(Z|D)$  is proportional to the likelihood term in eq. \ref{eq:logMAP}. We defer the proof to Appendix \ref{appendix:InfiniteLimitProof}%We will then show below that the resulting equation is tractable such that we can optimize it to obtain provably near-optimal MAP solutions:
%
%
\begin{theorem}  % using \phantom{invisible characters} to force horiz spacing
\label{theorem:infinitealpha}$\displaystyle \lim_{\alpha\rightarrow\infty} \big( \underset{Z}{\mathrm{argmax}} \ \log \text{P}(Z|D) \big)\\ \phantom{Theorem 1. lim}= \underset{Z}{\mathrm{argmax}} \sum_{d \in D} \sum^{|d|}_{i=1} \log Cat (w_{i,d} | z_{i,d}, \ \phi_{z_{i,d}})$ 
\end{theorem}
\vspace{-0.6em}
Our goal is now to describe a combinatorial objective function that admits a fast and near-optimal solution to the above expression under the explicit sparsity constraint that each document's words should be assigned to few topics on average. Absent this constraint, App.  \ref{appendix:expectednotopics} shows that sparsity (expected number of topics per document) is well approximated by $(\Phi/|D|)\sum_{d\in D}(1 - e^{-|d|/|\Phi|})$, yet we desire to explicitly compel sparse solutions. %First, observe that for a single document $d$, if we knew the subset $T_d \subseteq \Phi$ of candidate topics that were realized in $d$, then finding the MAP topic-word assignments, $\argmax_{Z_d} \text{P}(Z_d|d, T_d)$, is easy: just assign each word to the topic in $T_d$ with the highest probability of drawing it: $z_{d,i} = \argmax_{\tau \in T_d} \phi_\tau[w_{d,i}]$. Summing these, we have: 
First, observe that if we knew the subset $T_d \subseteq \Phi$ of candidate topics that were \mbox{realized} in a single document $d$, then finding document $d$'s MAP topic-word assignments, $\argmax_{Z_d} \text{P}(Z_d|d, T_d)$, is easy: just assign each word to the topic in $T_d$ with the highest probability of drawing it: $z_{d,i} = \argmax_{\tau \in T_d} \phi_\tau[w_{d,i}]$. Summing these gives $d$'s max conditional log posterior: 
\vspace{-1em}
\begin{align}
 \max_{Z_d} \big[ \log P(Z_d|T_d) \big] = \sum_{i=1}^{|d|} \ \max_{\tau\in T_d} (\log \phi_\tau[w_{d,i}])   
\label{eq:easy}
\vspace{-1em}
\end{align}
% It is clearly intractable to try all subsets $T_d$ for each document. However, we can approximate the solution we would obtain by doing so. The key idea is to formulate an objective function where the input is a set $E$ of \emph{links} (\emph{arrows in Fig. \ref{fig:mydiagrams}}) between topics and documents. We will say that the words in a document $d$ may only have been drawn from topic $\tau$ if our set of links $E$ includes a link between $\tau$ and $d$. Formally, consider the bipartite graph $G$ between topics $\tau \in \Phi$ and documents $d\in D$. Let $E$ denote a set of links $E:=\{\tau\Rightarrow d\}$, and let $E_d$ denote the set of topics with links to document $d$. Now we see that the solution to the expression in Theorem \ref{theorem:infinitealpha} is equivalent to maximizing the following set function:
It is intractable to try all subsets $T_d$ for each document to find the unconditional MAP $Z_d$, yet we can approximate this solution for \emph{all} documents. Consider a bipartite graph between topics $\tau \in \Phi$ and documents $d\in D$. Denote a set of links $E:=\{\tau\Rightarrow d\}$ 
(\emph{Figs. \ref{fig:diagiter1} \& \ref{fig:diagiter5} arrows}), 
and let $E_d$ denote topics with links to $d$. We will say that words in $d$ may only have been drawn from topic $\tau$ if $E$ includes a link from $\tau$ to $d$.  Now we see that the solution to the expression in Theorem \ref{theorem:infinitealpha} is equivalent to maximizing the following set function:
%
%
% \begin{align}
% \text{\emph{Objective function: } \ } &f(E) = \max \big[ \log P(Z|D,E) \big]  = \sum_{d\in D}\sum_{i=1}^{|d|} \ \max_{\tau\in E_d} (\log \phi_\tau[w_{d,i}])\\[-0.1em]
% \text{find \ } \underset{{\{\tau\Rightarrow d\}, \tau \in \Phi, d\in D}}{\argmax} &f(E) \text{\ subject to \ } |E|\le \kappa|D|; \ \ \ \ |E_d|\ge 1, \forall d
% \label{eq:ETM}
% \end{align}
%
%
\vspace{-0.2em}
\begin{align}
\vspace{-1.7em}
\label{eq:ETM}
\text{\emph{Objective function:} } f(E) &= \max \big[ \log P(Z|D,E) \big]  \\ & = \sum_{d\in D}\sum_{i=1}^{|d|} \ \max_{\tau\in E_d} (\log \phi_\tau[w_{d,i}])\notag\\[-0.1em]
\text{find } \underset{{\{\tau\Rightarrow d\}, \tau \in \Phi, d\in D}}{\argmax} f(E) &\text{\ s.t.}  \ |E|\le \kappa|D|; \  |E_d|\ge 1, \forall d \notag
\vspace{-5.5em}
\end{align}
%
%
%
%
% \footnote{Observe that $f$ captures the original LDA intuition that LDA can be viewed as a hierarchical extension of clustering (i.e. `mixture of unigrams') models \citep{blei2003latent}. In the case where the corpus $D$ contains just a single document, $f$ reduces to an objective
As LDA is a hierarchical extension of clustering \citep{blei2003latent}, note that $f$ is a hierarchical generalization of an objective that is well-studied in (exemplar) clustering, data summarization, and facility location \citep{gomes2010budgeted, lindgren2016leveraging, mirzasoleiman2015lazier,  pokutta2020unreasonable}.
By maximizing $f$ subject to the cardinality constraint $|E|\le \kappa|D|$, we seek a MAP solution where documents have varying counts of topics, but the \emph{average} number of topics linked to each document is bounded by $\kappa$. Thus, choosing a smaller $\kappa$ yields a sparser solution.
%seek to know the MAP topic assigned to each word in the data given that each document was drawn from no more than $\kappa$ topics \emph{on average}, Thus, choosing smaller $\kappa$ results in a sparser solution: documents will have varying numbers of topics, but the \emph{average} number of topics linked to each document will be no more than $\kappa$.
%This is practically convenient, as the user may provide an upper-bound (e.g. `the average document was drawn from no more than $4$ topics'), which is easy to obtain in practice via sampling.\footnote{We note that this sparsity constraint is \emph{not} a prior, as it constrains the realization of the draws. Rather, it gives a natural `realized sparsity' analogue to the standard LDA `underlying distribution sparsity' formulation.}

% We emphasize that this constraint only upper-bounds the maximum average number of topics per document; different documents will have different numbers of topics, and the actual average number of topics per document in the $Z$ we learn may be less\footnote{Note that due to the $\max$ function in \ref{eq:easy}, adding more topics to $\mathtt{E}$ that link to a particular document $d$ does not necessarily result in those topics being assigned to words in $d$. It is possible to add a link $[\tau\Rightarrow d]$ where each word in $d$ is already more likely to be drawn from one of the other topics linked to $d$ in $\mathtt{E}$.}  than $\kappa|D|$, as it will be guided by the likelihood in eq. \ref{eq:ETM}.

One challenge is that $f$ is undefined when some document has no links in $E$. We address this by extending the initialization technique of \citet{gomes2010budgeted} by linking each document to a single improper placeholder topic $p$ with infinitesimal probability on each word, then maximizing $\dot{f}$ (equivalent to maximizing $f$): $\dot{f}(E) = f(E \cup P) - f(P)$, where $P$ is the set of placeholder topics.

\textbf{Objective $\dot{f}(E)$ is a monotone submodular function.}
Observe that $\dot{f}$ is \emph{monotone}, as adding more links to $E$ cannot decrease $\dot{f}(E)$ due to the $\max$ function in $f$. $\dot{f}$ is also \emph{submodular} (see App. \ref{AppendixSubmodular}): Consider the \emph{\textbf{marginal value}} $\dot{f}_{E}(A) \coloneqq \dot{f}(E \cup A) - \dot{f}(E)$ of adding links $A = \{[\tau$$\Rightarrow$$d]\}$ to $E$. As the set of topics linked to $d$ in $E$ grows, any topic $\tau\in A$ will be the maximum-probability topic for (weakly) fewer words in $d$, so it will contribute fewer terms to $\dot{f}$, and the marginal value of adding it to $E$ is weakly decreasing:  $\dot{f}_{E}(A) \geq \dot{f}_{E'}(A), \forall \tau \in A \subseteq \Phi, d\in D,  E \subseteq E'$.

\section{Greedy guarantees a near-optimal global \& per-doc solution but is practically infeasible}
\label{sec:submodular}

Consider the simple Greedy algorithm that iteratively adds to $E$ the link with the greatest marginal value. Because $\dot{f}$ is monotone submodular, it is well-known that
% \textbf{A near-optimal $1-1/e$ approximation to the optimal value of $\dot{f}$.} 
Greedy returns a list of topic-document links $E$, where \textbf{the first $j$ links in $E$ are a near-optimal $1-1/e$ approximation to the MAP topic-word assignments s.t. $|E|\le j$.} Thus, one chooses sparsity constraint $\kappa|D|$ as an upper bound, then \emph{a single run} of Greedy gives near-optimal solutions for any number of topics from an average of $1$ to $\kappa$ per document.
%\textbf{Every document's topics are near-optimally chosen.} 
Also, as $\dot{f}$ is a sum of (per-document) monotone submodular functions, \textbf{the topics $E_d$ that Greedy assigns to \emph{each document} $d$ obtain a $1-1/e$ approximation to $d$'s optimal MAP value for any $|E_d|$ topics} (see App. \ref{app:approximationguarantee}). Unfortunately, Greedy is practically infeasible, as each of its $\kappa|D|$  iterations is %$O(\ell\kappa|\Phi||D|^2)$, where $\ell$ is the max document length (see App.  \ref{app:simplegreedy}). 
$O(|d|\kappa|\Phi||D|^2)$ (see App.  \ref{app:simplegreedy}).

\vspace{-0.3em}
\section{Fast practical algorithms for big data}
\label{sec:fastalgorithms}
We now describe how to leverage LDA's independence assumptions to obtain serial and parallel algorithms that \emph{find MAP solutions with these same near-optimal guarantees faster than mainstream LDA algorithms find locally modal solutions.} Our algorithms solve some of the largest submodular optimization problems studied anywhere in terms of the size of the ground set and solution set (see App. \ref{app:experiments}). We call our approach Exemplar-LDA (E-LDA) to reflect the inspiration of exemplar data summarization. 

We now describe \textsc{FastGreedy-E-LDA}, \textsc{FastInitialize}, and the subroutine \textsc{UpdateHeap}. \textsc{FastGreedy-E-LDA} computes the ordered set $E$ of near optimal topics assigned to documents, which implies (via eq. \ref{eq:easy}) the near-optimal topic $z_{d,i}$ from which each word was drawn. It inherits the deterministic near-optimal guarantees of Greedy, but its iterations are exponentially faster.

In brief, the key idea that enables each speedup is that the marginal value (increase in the log probability objective $\dot{f}$) of one link to document $d$ is independent of everything except the current best topic associated with each word in $d$. With suitable memoizations and a heap data structure, we replace an $O(|\Phi||D|^2)$ runtime factor with an $O(\log|D|)$ factor. We can also initialize each document with its best link. App.  \ref{app:fastgreedy} details our strategies for leveraging LDA's structure to obtain fast convergence.

 In this algorithm, $\mathbf{P}_{|D|\times |V|}$ memoizes the log probability of each word in the data per the current iteration's topic-word assignments $Z$; $\mathbf{p}_{1 \times |D| }$ memoizes the current summed log probability of each document per the current $Z$; $\mathbf{M}_{|D|\times|\Phi|}$ memoizes all current marginal values of links; and $\mathbb{m}$ memoizes each document's best not-yet-added topic and its marginal value in a max-heap, so the best link can be accessed
 in $O(\log|D|).$ We denote by $\odot$ the Hadamard product, and by $\mathbb{1}$ a matrix of $1$'s, and we denote all entries corresponding to document $d$ by e.g. $[d,:]$. We denote by \textsc{Emax}$(a,b)$ the elementwise maximum of $a$ and $b$.

%%%%%%%%%%%%%%%%%%%%%%   FAST INITIALIZATION OF FIRST TOPIC PER DOCUMENT  %%%%%%%%%%%%%%%%%%%%%%
\vspace{-0.3em}

\FloatBarrier  % prevent earlier floats from jumping in front

\begin{algorithm}[H]
\caption*{\textbf{FastInitialize}: \emph{Compute each document's first linked topic.}}
\addcontentsline{loa}{algorithm}{FastInitialize}

\begin{algorithmic}
    \INPUT  Doc-word matrix $\mathbf{D}_{|D|\times|V|}$, topic matrix $\mathbf{\Phi}_{|\Phi| \times |V|}$ \\
    %
    % INITIALIZE
    \STATE  $\mathbf{P}_{|D|\times |V|} \gets \log(\epsilon)\cdot \mathbf{D}$ \\
    \STATE  $\mathbf{p}_{1 \times |D|} \gets \mathbb{1}_{1 \times |V|} \cdot \mathbf{P}^T  $ \\
    \STATE  $\mathbf{M}_{|D|\times|\Phi|}\gets \mathbf{D} \cdot \mathbf{\Phi}^T - \mathbf{P} \cdot \mathbb{1}_{|V|\times |\Phi|}$
    \STATE  $E \gets$ the $|D|$ links $[\tau^*_{d^*=1}\text{$\Rightarrow$} 1], \dots, [\tau^*_{d^*=|D|}\text{$\Rightarrow$} |D|]$ \\ \hphantom{E arro}corresp. to the max value per row of $\mathbf{M}$\\
    \STATE  \textbf{for} link $[t^*\text{$\Rightarrow$} d^*] \in E$ \textbf{do}:
    \STATE  \ \ \ \ \ $\mathbf{P}, \mathbf{p}, \mathbf{M}, \mathbb{m} \gets $\textsc{Update($[\tau^*\text{$\Rightarrow$} d^*], \mathbf{D}, \mathbf{\Phi}, \mathbf{P}, \mathbf{p}, \mathbf{M}, \mathbb{m}$})
            \STATE \textbf{return} \ $E, \mathbf{P}, \mathbf{p}, \mathbf{M}, \mathbb{m}$%,\ \  $Z$ 
  \end{algorithmic}
  \label{alg:FastInit}
\end{algorithm}

%\textcolor{red}{Where $\phi_{phantom}$ is an improper topic with constant small probability on all words, e.g. $[1^{-10}]^{|V|}$}

%%%%%%%%%%%%%%%%%%%%%%   FastGreedy-E-LDA  %%%%%%%%%%%%%%%%%%%%%%
\vspace{-0.8em}

%\vspace{-1em}
\begin{algorithm}[H]
\caption*{\textbf{FastGreedy-E-LDA}: \emph{Main routine.}}
\addcontentsline{loa}{algorithm}{FastGreedy-E-LDA}

\begin{algorithmic}
    	\INPUT  Doc-word matrix $\mathbf{D}_{|D|\times|V|}$, topic matrix $\mathbf{\Phi}_{|\Phi| \times |V|}$, \\ \hphantom{puttt}sparsity upper-bound $\kappa$ \\
     \STATE $E, \mathbf{P}, \mathbf{p}, \mathbf{M}, \mathbb{m} \gets$ \textsc{ FastInitialize}$(\mathbf{D}, \mathbf{\Phi})$\\
     %
     % % RUN THE FOR LOOP
        \STATE \textbf{for} $(\kappa-1)|D|$ steps \textbf{do}\\
           %
           % Add the best topic-doc link to S
           \STATE \ \ \ \ \ $t^*, d^* \gets \text{ the topic and doc. corresp. to \emph{extract }} \max (\mathbb{m})$ \\
           \STATE \ \ \ \  $E \gets [\tau^*\text{$\Rightarrow$} d^*] $ \\
    \STATE  \ \ \ \  $\mathbf{P}, \mathbf{p}, \mathbf{M}, \mathbb{m} \gets $\textsc{Update($[\tau^*\text{$\Rightarrow$} d^*], \mathbf{D}, \mathbf{\Phi}, \mathbf{P}, \mathbf{p}, \mathbf{M}, \mathbb{m}$})
           \STATE \textbf{return} \ $E $%,\ \  $Z$ 
\end{algorithmic}
\label{alg:AccelGreedy}
\end{algorithm}
%$\Phi_R$ stands for realized; X is the random set;
%Where $\mathbb{1}$ is a column vector of $1$'s
%$\odot$ denotes the Hadamard (elementwise) product;

%%%%%%%%%%%%%%%%%%%%%%   Fast memo update   %%%%%%%%%%%%%%%%%%%%%%
\vspace{-0.8em}
\begin{algorithm}[H]
\caption*{\textbf{Update}: \emph{Update the heap  with the current log prob. of $d$ and each word in $d$ after adding a new link to the solution, then update marginal values of each link that includes $d$.}}
\addcontentsline{loa}{algorithm}{UpdateHeap subroutine}
% \textbf{update} log probs. of current $Z$ and marginal values for links to $d$:\\
\begin{algorithmic}
    % \INPUT  Doc-word matrix $\mathbf{D}_{|D|\times|V|}$, topic matrix $\mathbf{\Phi}_{|\Phi| \times |V|}$ \\
    \INPUT  Doc-word matrix $\mathbf{D}_{|D|\times|V|}$, topic matrix $\mathbf{\Phi}_{|\Phi| \times |V|}$, \\ \hphantom{puttt}link $[\tau^*\text{$\Rightarrow$} d^*]$, storage  $\mathbf{P}, \mathbf{p}, \mathbf{M}, \mathbb{m}$\\
    \STATE  $\mathbf{p}[d^*] \gets \mathbf{p}[d^*] + \mathbf{M}[d^*,t^*]$ \\%\textcolor{red}{\text{ Maybe remove this line and just do a prod with a $\mathbf{1}$ when we use \textbf{p} in \textbf{M} line below}}\\
    %
    % Now a loop to update the marginal values for each doc
        \STATE  $\mathbf{P}[d^*,:] \gets \textsc{Emax}( \mathbf{P}[d^*,:], \ \ \  \mathbf{\Phi}[t^*,:] \odot \mathbf{D}[d^*,:]) $ \\
        \STATE $\mathbf{\sigma}_{d^*} \gets \mathbf{\Phi} \odot [\mathbf{D}[d^*,:];\dots;\mathbf{D}[d^*,:] ])$\\
        \STATE $\mathbf{M}[d^*,:] \gets  \mathbb{1}_{|V|\times 1} \cdot \textsc{Emax} ( \mathbf{P}[d^*,:] , \ \  \mathbf{\sigma}_{d^*}) - \mathbf{p}[d^*] $\\
        \STATE $\mathbb{m} \gets insert \max \mathbf{M}[d^*,:]$
        %
   % \STATE $ \mathbb{m}:\mathbb{m}[d^*] \gets insert \max(\mathbf{M}[d^*,:]) $\\
            \STATE \textbf{return} \ $\mathbf{P}, \mathbf{p}, \mathbf{M}, \mathbb{m}$ 
  \end{algorithmic}
  \label{alg:memo}
\end{algorithm}

\FloatBarrier

%We now derive the complexity of the vanilla \textsc{Simple-Greedy} algorithm described in Section \ref{ssec:greedy} and 
%compare it to the \textsc{FastGreedy-E-LDA} algorithm described here:

%%Simple-Greedy completes in $\kappa|D|$ iterations. Each iteration $i$ of Simple-Greedy computes the marginal value of $|\Phi||D| - i$ links. Without the speedups described in \textsc{FastGreedy-E-LDA}, computing one of these marginal values entails solving $f(E^i \cup [\tau\Rightarrow d]) - f(E^i)$, where $[\tau\Rightarrow d]$ is a new topic-doc link (where $E^i$ denotes the solution at iteration $i$). Computing a value of $f$ requires, for each document in $D$, finding the max log probability of each word in $d$ among $E_d$ topics (or $E_d+1$ topics for $f(E^i \cup [\tau\Rightarrow d])$, then summing the result. Denote the number of words in the longest document by $\ell$. Thus each marginal value requires $O(\ell|E_d||D|)$ which is $O(\ell\kappa|D|)$. An iteration of Simple-Greedy requires computing $O(|\Phi||D|)$ such marginal values, so each iteration is $O(\ell\kappa|\Phi||D|^2)$. 

% We state our guarantees formally here in Theorem \ref{theorem:acceleratedgreedy}, and give the complexity analysis below:

\newpage           %%%%% NEWPAGE
\begin{theorem}
\label{theorem:acceleratedgreedy} 
\textsc{FastGreedy-E-LDA} returns a solution $E$ in $\kappa|D|$ iterations each of complexity $O(\log |D| + \ell|\Phi|)$. In the worst-case, $f(E)$ is a $1-1/e$ approximation to the log posterior value of the (optimal) LDA MAP topic-word assignments (eq. \ref{eq:ETM}) subject to the user-chosen sparsity (cardinality) constraint $|E|\le \kappa |D|$; \ \ The value $f(E[{:}{j}])$ of the first $j$ links in $E$ is a $1-1/e$ approximation to the log posterior value of the optimal LDA log MAP topic-word assignments, eq. \ref{eq:ETM}, subject to $|E|\le j$;\ \  The topics $E_d$ connected to each document $d$ obtain a log posterior value  $\sum_{i=1}^{|d|} \ \max_{\tau\in E_d} (\log \phi_\tau[w_{d,i}]) $ that is a $1-1/e$ approximation to the log posterior value of the single-document MAP topic-word assignments for document $d$ that contain no more than $|E_d|$ topics; \ \ The first $j$ links in $E_d$ give a $1-1/e$ approximation to the log posterior value of the MAP topic-word assignments for $d$ subject to $|E_d|\le j$. \emph{We defer the proof to App.  \ref{app:fastgreedy}.}
\end{theorem}
%     }
% }

% %%%%%%%%%%%%%%
% \section{TWO MAP topic assignment for each word in the data}
% \label{sec:myfunctions}

% Suppose a finite set containing all topics $\Phi$ is known. For example, the topics may have already inferred by another algorithm \citep{arora2013practical}, or we may simply know that the topics that were realized in the generative model (i.e. the subset of topics that were actually drawn while generating the data) are contained in a large candidate set (see Section \ref{sec:candidates} below).

% In this case, after integrating out $\Theta$ per eq. \ref{eq:LDAmarginal}, and ignoring the nuisance document length term (which is standard \citep{blei2003latent}), we arrive at the canonical MAP topic assignment inference problem first studied in seminal work by \citet{sontag2011complexity}:

% \begin{align}
%     \text{P}(Z | D) \propto
%         \displaystyle \prod_{d \in D} \Big(
%         \frac{\Gamma(\prod^k_{\tau=1} \alpha_\tau)\prod^k_{\tau=1}\Gamma(n_{d,\tau} + \alpha_\tau)}{\Gamma(\prod^k_{\tau=1}\alpha_\tau + N) \prod^k_{\tau=1} \Gamma(\alpha_\tau)}
%           \prod^{|d|}_{i=1} Cat (w_{d,i} | z_{d,i}, \ \phi_{z_{d,i}})   \Big)

\medskip
\section{Near optimal solutions in log parallel time}
%\section{Near optimal solutions in logarithmic parallel time}
\label{sec:FASTalgorithm}
% Recall that the parallel runtime of optimization is measured by \emph{adaptivity}, which is the number of sequential rounds an algorithm requires when polynomially many queries ($\dot{f}$ evaluations) can be executed in parallel in each round.
Because we have proven that $\dot{f}$ is monotone submodular, we could in theory use the recent breakthrough in low-adaptivity parallel algorithms to obtain near-optimal solutions in logarithmic parallel time (adaptivity) \citep{balkanski2018adaptive, balkanski2019optimal, breuer2020fast, chekuri2018submodular, ene2019, FMZ19}.
%BS18b,BBS18,EN19,FMZ19,FMZ19b,KMZLK19,
%chekuri2018matroid,chekuri2018submodular,BRS19,%balkanski2019optimal,
%ene2019,chen2019,esfandiari2019adaptivity,QS19
The challenge is that simulating a single query for $\dot{f}$ is  $O(\ell \kappa|\Phi||D|^2)$ in a vanilla implementation, yielding infeasible runtimes in practice. 
 By extending our techniques, we show we can in fact simulate each query in time that is both practically fast and independent of the size of the data $|D|$, enabling fast parallel solutions in practice. We focus on the \textsc{FAST} algorithm \citep{breuer2020fast} because it is practical and because it evaluates one link at a time, not sets of links, so our techniques above apply. We defer the full algorithm and analysis to App.  \ref{FAST_MODIFIED} and \ref{app:FASTalgorithmquerysimulation}.
% \section{Near optimal solutions in logarithmic parallel time}
% \label{sec:FASTalgorithm}
% % Recall that the parallel runtime of optimization is measured by \emph{adaptivity}, which is the number of sequential rounds an algorithm requires when polynomially many queries ($\dot{f}$ evaluations) can be executed in parallel in each round.
% Very recently, the breakthrough in low-adaptivity optimization yielded parallel algorithms that near optimally maximize any monotone submodular function w.p. in logarithmic parallel time (adaptivity) \citep{balkanski2018adaptive, balkanski2019optimal, breuer2020fast, chekuri2018submodular, ene2019, FMZ19}.
% %BS18b,BBS18,EN19,FMZ19,FMZ19b,KMZLK19,
% %chekuri2018matroid,chekuri2018submodular,BRS19,%balkanski2019optimal,
% %ene2019,chen2019,esfandiari2019adaptivity,QS19
% However, for our objective $\dot{f}$, simulating a single query is  $O(\ell \kappa|\Phi||D|^2)$ in a vanilla implementation, yielding practically infeasible runtimes. 
%  By extending our techniques, we show we can simulate each query in time that is both practically fast and independent of the size of the data $|D|$, enabling fast parallel solutions in practice. We focus on the \textsc{FAST} algorithm \citep{breuer2020fast} because it is practically fast and because it evaluates one edge at a time, not sets of edges, so our techniques above apply. We defer the algorithm and proofs to Appendix \textbf{\textcolor{red}{STUFF REFERENCES}}.
% % \citet{breuer2020fast} proves the following theorem (changed to our notation):
\bigskip
\begin{theorem} \citet{breuer2020fast}.
\label{theorem:FASTguarantees} For 
 $\kappa|D| \geq \frac{2 \log(2\delta^{-1} \ell)}{\varepsilon^2 (1 - 5\varepsilon)}$ and $\varepsilon \in (0, 0.1)$, where  $\ell = \log(\frac{\log \kappa|D|}{\varepsilon})$, \textsc{Fast} \ with sample complexity $m = \frac{2 + \varepsilon}{\varepsilon^2(1 -  3\varepsilon)} \log(\frac{4\ell\log |\Phi||D|}{\delta \varepsilon^2})$ has at most $\varepsilon^{-2} \log(|\Phi||D|)  \ell^2$  adaptive rounds,   
$2\varepsilon^{-2}  \ell |\Phi||D| +  \varepsilon^{-4} \log(|\Phi||D|) \ell^2 m$ queries, and achieves a $1 - \frac{1}{e} - 4 \varepsilon$ approximation 
to the log MAP topic-word assignments (eq. \ref{eq:ETM}) s.t. sparsity constraint $|E|\le \kappa|D|$ with prob. $1 - \delta$.
\end{theorem}
\bigskip
\begin{theorem} 
\label{theorem:FASTcomplexityofsim} For the MAP objective $\dot{f}$, each query in \textsc{Fast} can be simulated in runtime that is either $O(\ell)$ or $O(\ell|\Phi|)$ with suitable memoization. \emph{We defer analysis to App. \ref{app:FASTalgorithmquerysimulation}.}
\end{theorem}

\newpage
\section{Exemplar-LDA: Learning LDA on a candidate set of interpretable topics}
\label{sec:candidates}
While we can obtain the candidate set $\Phi$ of topics from existing algorithms (see \citet{arora2018learning}), this does not address the interpretability problems described above.  We now show that our algorithms enable a new way to solve topic models by learning LDA from the large candidate set of all topics that meet standard interpretability criteria. Specifically, we generate interpretable candidate topics via coherence functions that are already widely used to assess topic quality post hoc. For example, \textsc{UMass} coherence  \citep{mimno2011optimizing} measures the raw similarity $s$ between a word-of-interest $w^*$ and any other word $v\in V$ as the ratio of documents $D_{w^*}$ that contain $w^*$ to documents  $D_{w^*,v}$ that contain both $w^*$ and $v$. $s$:$\ s(w^*,v)^{\textsc{UMass}}= \frac{|D_{w^*,v}| + \epsilon}{|D_{w^*}|}$. Any such similarity function can be used to define a \emph{new} topic `about' a word of interest $w^*$ by ascribing distributional mass to each word $v\in V$ in proportion to $s(w^*,v)$. For example, to generate a new topic $\phi_{w^*=\text{`physics'}}$ about the word of interest `physics', we set $\phi[v] \propto s(\text{`physics'},v)$ for each word $v$ in the vocabulary. A topic generated in this way has a  formal interpretation, as it ascribes mass per a simple function that is known to the researcher. At a high level, such a topic is `rigorously labeled' by its word. To obtain a candidate set $\Phi$ of topics, we simply generate a topic for every word $w^*\in V$ (or for every phrase e.g. $w^*\in (V\cup 2$-grams $\cup$ 3-grams$)$). \emph{We emphasize that we don't require all topics to be good, but merely that the topics we want to learn are included in this large candidate set.} We consider three topic generating functions:

%\footnote{We briefly note that this approach also provides a means to alleviate inferential challenges specific to `small data/small corpus' topic modeling. Specifically, it has become standard for researchers who wish to measure the `external validity' of the topics returned for their corpus to compute their topics' coherence scores on an external reference corpus such as Wikipedia \citep{newman2010automatic, stevens2012exploring, roder2015exploring}. Likewise, using the techniques described above, it is possible to \emph{generate} our set of topics from e.g. Wikipedia, then perform MAP inference on our corpus by selecting from this `externally valid' set.}

%\footnote{For example, to generate a new topic $\phi_{w^*=\text{'physics'}}$ about the word of interest `physics', we set $\phi[v] \propto s(\text{`physics'},v)$ for each word $v$ in the vocabulary.}

% Thus, $s(w^*,v)^{\textsc{UMass}}$ measures the fraction of documents containing a word of interest that also contain word $v$. 

% his set can be obtained from existing algorithms, though this does not address all of the interpretability problems described above. Therefore, our third challenge is to show that it is possible to \emph{efficiently generate} this comprehensive candidate set of topics via same standard coherence functions that are widely used to \emph{measure} topics' interpretive quality ex-post. 

%\smallskip
\textbf{UMass:} $\phi_{w^*}[v] \propto s(w^*,v)^{\textsc{UMass}}$. This ironically yields low-coherence solutions---see Section \ref{sec:experiments} and App. \ref{app:umassbad}.
%achieve poor \textsc{UMass} coherence scores post hoc (see Fig. \ref{fig:manybenchmarks}), as it generates insufficiently sparse topics that all assign mass too uniformly across the vocabulary to perform well. 

%\smallskip
\textbf{Exp-UMass:} $\phi_{w^*}[v] \propto  \exp( s(w^*,v)^{\textsc{UMass}} )$. This approach generates high quality sparse topics.% and lets us optimize $\dot{f}$ on the raw word co-occurrence counts (the $log(\phi)$ in $\dot{f}$ cancels the $\exp$).

%\smallskip
\textbf{Co-occurrence:} In exemplar clustering and data summarization, it is standard to measure points' similarity via their features' dot product. We extend this to LDA, and the result is highly performant. In LDA `points' are documents, `features' are words, and the dot product counts word co-occurrences.  So, we generate topics via:
$\phi^\text{co-occurrence}_{w^*}[v] \propto \exp(|D_{w^*,v}|+\epsilon)$. Recall that Theorem \ref{theorem:infinitealpha} had all $\alpha$ priors equal for simplicity, yielding $\theta[\tau] \rightarrow 1/|\Phi|, \forall \tau$. Instead, consider the case where the probability of drawing the topic associated with word $w^*$ is proportional to $w^*$'s overall popularity (total co-occurrences): if $\phi^\text{co-occurrence}_{w^*}$ is the $\tau$'th topic, then $\theta[\tau] \propto \sum_{v\in V} \exp(|D_{w^*,v}|+\epsilon).$ Now, $f$ simplifies to a (hierarchical extension of a) familiar data summarization objective with topics as the `exemplars', and we optimize on the raw dot products (App.  \ref{cooccurrence_proof} gives details):
\vspace{-1em}
\begin{align}
f(\mathtt{E}) \text{$=$} \max \big[ \log P(Z|D,\mathtt{E}) \big] \text{$=$} \sum_{d\in D}\sum_{i=1}^{|d|}  \max_{\tau\in \mathtt{E}_d} (|D_{w^*,w_{d,i}}|) \notag % REMOVING LABEL TO SAVE A LINE
%\label{eq:cooccurrenceobjective}
\vspace{-1em}
\end{align}
%, where $w^*$ is the word used to generate topic $\tau$.

% If a word has high co-occurrence with the other words in (part of a) a document, then it is intuitively representative of that (part of the) document. We now show that it is possible to define our set of topics to capture this intuition such that when we optimize E-LDA, we are just optimizing on the raw dot product of the (transposed) binary document-word matrix and itself. \emph{This naturally results in LDA solutions where each document is well-represented by the keywords $(w^*)$ associated with its assigned topics.}

%\newpage

\section{Near-optimal out-of-sample inference \& downstream causal inference: }
\label{sec:causalneurips}
We now highlight an important application for our results. There is now great excitement in the social sciences about using LDA topics as treatments for downstream causal inference frameworks that infer how topics cause outcomes. A key challenge is that these frameworks require (1) that treatments (topics) are independent of topic assignments, and (2) that each document's causal response is independent of other documents' assigned topics (SUTVA). To enforce this, the gold-standard approach splits documents into training/test sets, learns a topic model on the training set, then infers test set documents' topics using (separate) out-of-sample algorithms \cite{egami2022make}. However, even if the training set LDA model is optimally fit, standard out-of-sample algorithms used for the test set have no guarantees. They are also too complex to manually vet, and they are known to exhibit bias and sensitivity to random initialization \citep{wallach2009evaluation, egami2022make}. This results in spurious and/or biased causal inferences.

In contrast, we can modify $\dot{f}$  to enforce the required independences for the out-of-sample test set inference. We then obtain the near-optimal $1-1/e$ approximation to each document's $d$'s log MAP assignments in $\kappa_d$ iterations each of complexity that is just $O(\ell|\Phi|)$. To do this, we solve:
\begingroup
% Change spacing definitions only within this group
%\thinmuskip=3mu
%\medmuskip=4mu plus 2mu minus 4mu
%\thickmuskip=5mu plus 5mu 
\thinmuskip=3mu minus 10mu
\medmuskip=4mu plus 1mu minus 10mu
\thickmuskip=5mu plus 1mu minus 10mu
\vspace{-.1em}
\begin{align}
%\begin{align}
&\text{\emph{Out-of-sample objective $f^+$}: \ } \notag \\&f^+(E_d)\text{$=$}\max \big[ \log P(Z_d|d,E_d) \big]  \text{$=$} \sum_{i=1}^{|d|} \ \max_{\tau\in E_d} (\log \phi_\tau[w_{d,i}])\notag\\[-0.5em]
&\text{find \ } \underset{{\{\tau\Rightarrow d\}, \tau \in \Phi}}{\argmax} f^+(E_d) \text{\ s.t. \ } |E_d|\le \kappa_d; \  |E_d|\ge 1 
\label{eq:ETMOOS}
\end{align}
\endgroup
Appendices \ref{sec:causalbackground} \& \ref{sec:causal} give details. The resulting MAP assignments are well-defined as treatments (unlike documents' underlying topic mixtures $\theta_d$). This approach also conveys the interpretability advantages of our topic generators, which are paramount in the causal inference setting.

%\newpage

\section{Experiments}
\label{sec:experiments}

% Our experiments show that our algorithm outperforms the current standard for d
% social science applications of data exploration and causal inference in terms of the two metrics that these researchers care about: log probability and coherence.

Our goal in this section is to show that beyond its provable guarantees and interpretability properties, E-LDA outperforms baselines in the two dimensions that social science researchers prioritize: posterior probability and topic coherence. We conduct two sets of experiments. In the first set, we solve LDA models using the main variations of the MALLET Gibbs LDA baseline \citep{mccallum2002mallet} that is standard in social science. We test whether our algorithms obtain better topic-word assignments in terms of posterior probability \emph{even when constrained to use the Gibbs topics and keep the same avg. sparsity as the Gibbs assignments.}

% In the second set of experiments, we compare the semantic quality of baselines' topics (incl. recent neural and LLM topic models) to the topics that E-LDA learns when its candidate set is initialized by our topic generators. We use the standard UMass coherence measure of topic quality \citep{mimno2011optimizing, roder2015exploring}, which tests the (log) probability that the top (i.e. highest-probability)  words in a topic co-appear in documents. Coherence has been shown to correlate with human judgments of semantic quality. App. \ref{app:experiments} gives details.

In the second set of experiments, we add neural and LLM/BERT-based topic model baselines, and we compare the semantic quality of baselines' topics to the topics that E-LDA learns when its candidate set is initialized by our topic generators. %We use the standard UMass coherence measure of topic quality, which correlates with human judgments of semantic quality \citep{mimno2011optimizing, roder2015exploring}. 
We measure topics' semantic quality via standard UMass coherence, which correlates with human quality judgments \citep{mimno2011optimizing, roder2015exploring}.

\textbf{Code.} Python code \& data (see App. \ref{app:experiments}) can be found at: \textbf{\url{https://github.com/BreuerLabs/E-LDA} }

\textbf{Baselines.} We consider three LDA baseline algorithms, each run with $k=100$ topics and averaged over $10$ runs per dataset: \emph{(1) Gibbs} -- The MALLET Gibbs LDA solver with the usual uninformative prior $\alpha$=$1$; \ \emph{(2) Gibbs-Bayes (G-B)} -- Gibbs with Bayesian $\alpha_\tau$ tuning; \  \emph{(3) Gibbs-Sparse (G-S)} -- A Gibbs solver with $\alpha$=$0.1$. For the log posterior experiments, we also give a \emph{rand}$\kappa$ baseline equal to the log posterior of the Gibbs assignments after randomly permuting topics (mean of $30$ permutations/run). For topic coherence experiments (Experiments Set $2$), we add recent neural and LLM-based baselines: \emph{BERTopic} \citep{grootendorst2022bertopic}, \emph{AVITM} \citep{srivastava2022autoencoding},  \emph{ETM}  \citep{dieng2020topic}, and \emph{FASTopic} \citep{wu2024fastopic}.

\textbf{Datasets.} We consider three well-studied social science datasets: Reuters news articles \citep{lewis1997reuters}, US Congressional speeches \citep{thomas2006get}, and social media posts from political news discussion boards \citep{lang1995newsweeder}.

% \paragraph{Evaluating topics' semantic quality.} Specifically, for a given topic $\tau$, the normalized coherence is $C(\tau,h^*)= \frac{2}{h^*(h^*-1)}\sum_{h=2}^{h^*} \sum_{\ell=1}^{h-1} \log s(\tau_h,\tau_\ell)$, where $\tau_h$ and $\tau_{ell}$ are the $h$'th and $\ell'th$ most probable words, resp., according to topic $\tau$.

% \input{SEC_TABLE_OBJVALS_CONVENTIONAL_LDA_MAPNOCONSTANT}

% \input{SEC_TABLE_OBJVALS_CONVENTIONAL_LDA_MAPWITHCONST}

\textbf{Experiments Set 1 Results.} Table \ref{tab:OBJval}  and App. \ref{app:experiments} give the log posterior values  of the assignments ($\dot{f}$ values). Our algorithms obtain uniformly higher probability assignments ($p < 10^{-10}$) on all $90$ experiments ($3$ datasets$\times$$3$ baselines$\times$$10$ runs). E-LDA obtains these results despite using baselines' topics and constraining its solution to the same assignment sparsity chosen by baselines' algorithms.

\begin{table}[h]
\centering
\tabcolsep=2.4pt
\resizebox{\columnwidth}{!}{%
\begin{tabular}{@{}|l|lll|lll|lll|@{}}
    % \begin{tabular}{|r|p{2em}|p{3em}|p{3em}|r|p{2em}|p{2.5em}|p{2.5em}|}
\toprule
           & Gibbs & E-LDA & \color{gray}{rand$\kappa$} & G-B & E-LDA & \color{gray}{rand$\kappa$} & G-S & E-LDA & \color{gray}{rand$\kappa$} \\ \midrule \midrule
Congress   &   -2.05    &    \textbf{-1.78}      &     \color{gray}{-10.92}       &  -2.47     &      \textbf{-2.22}        &  \color{gray}{-10.22}    &    -2.43              & \textbf{-2.14}  &  \color{gray}{-10.31}   \\ \midrule
Reuters   &   -2.80    &   \textbf{-2.52}    &   \color{gray}{-16.54}   &      -3.66      &   \textbf{-3.43}    &  \color{gray}{-15.19}    &      -3.49       &    \textbf{-3.13}   &   \color{gray}{-15.41}   \\ \midrule
NewsGroups &     -1.57    &    \textbf{-1.40}      &     \color{gray}{-6.71}       &  -1.70     &      \textbf{-1.61}        &  \color{gray}{-6.57}    &    -1.68              & \textbf{-1.55}  &  \color{gray}{-6.52}   \\ \bottomrule
\end{tabular}
} % END RESIZEBOX
\caption{Log posterior of assignments ($\dot{f}$ value) with same topics \& sparsity (10 run avg., values $\times10^6$, see App. \ref{app:experiments}).}
\label{tab:OBJval}
\end{table}

\textbf{Runtimes.} The research-grade Python codes of our non-parallel algorithm take $2$-$3$ sec., approximately the same time it takes for the highly optimized MALLET Java Gibbs algorithm to run just $1$-$2$ of its $\sim$$8$ outer loop iterations.

% \begin{figure}
% \centering
% \begin{subfigure}[t]{1.0\textwidth}
% \includegraphics[width=\textwidth]{coherence_figures/BENCHMARK_plots_MAINSIMPLEerrorbar_eldak_noGoogle.pdf}
%    \caption[E-LDA coherence vs. state-of-the-art benchmark]{Mean topic coherence of E-LDA \textsc{co-occurrence} and \textsc{exp-UMass} solutions and Gibbs solutions, computed based on the top $M=5, 10, \dots, 25$ most popular words in each topic. Unlike E-LDA, Gibbs solution quality varies from one run to the next, so error bars plot the best and worst mean coherence over $10$ re-runs to show the range of typical performance.}
%    \label{fig:coherence_main} 
% \end{subfigure}
% \begin{subfigure}[b]{1.0\textwidth}
% \includegraphics[width=1.0\textwidth]{coherence_figures/BENCHMARK_plots_MAINSIMPLEWITHBOX_eldak_noGoogle.pdf}
%    \caption[Detail of E-LDA coherence vs. state-of-the-art benchmark]{Distribution of individual topic coherences of E-LDA and Gibbs: Means denoted by points and best/worst denoted by error bars.}
%    \label{fig:coherence_box}
% \end{subfigure}
% \caption[E-LDA coherence vs. state-of-the-art benchmark]{E-LDA coherence vs. state-of-the-art Gibbs benchmark.}
% %\label{fig:convergence}
% \end{figure}

\begin{figure*}[h] % THE STAR MAKES IT ACROSS TWO COLUMNS

\centering
\begin{subfigure}[b]{0.92\textwidth}
\includegraphics[width=1.0\textwidth]{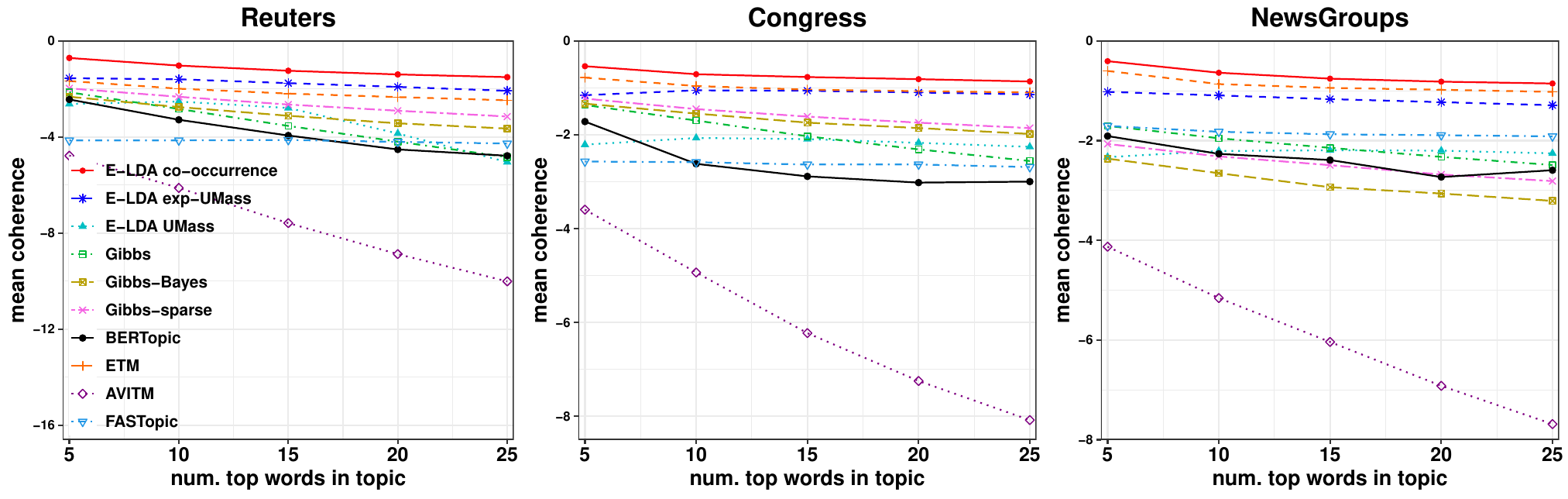}
   %\caption{Mean topic coherence of E-LDA \textsc{co-occurrence} and \textsc{exp-UMass} solutions and Gibbs solutions, computed based on the top $M=5, 10, \dots, 25$ most popular words in each topic. Unlike E-LDA, Gibbs solution quality varies from one run to the next, so error bars plot the best and worst mean coherence over $10$ re-runs to show the range of typical performance.}
   \caption[E-LDA coherence vs. baselines]{\textbf{Experiments Set 2:} E-LDA topic coherence vs. baselines for various num. top words $h^*$ (10 run avg.).}
   \label{fig:manybenchmarks} 
\end{subfigure}
\begin{subfigure}[b]{0.92\textwidth}
\includegraphics[width=1.0\textwidth]{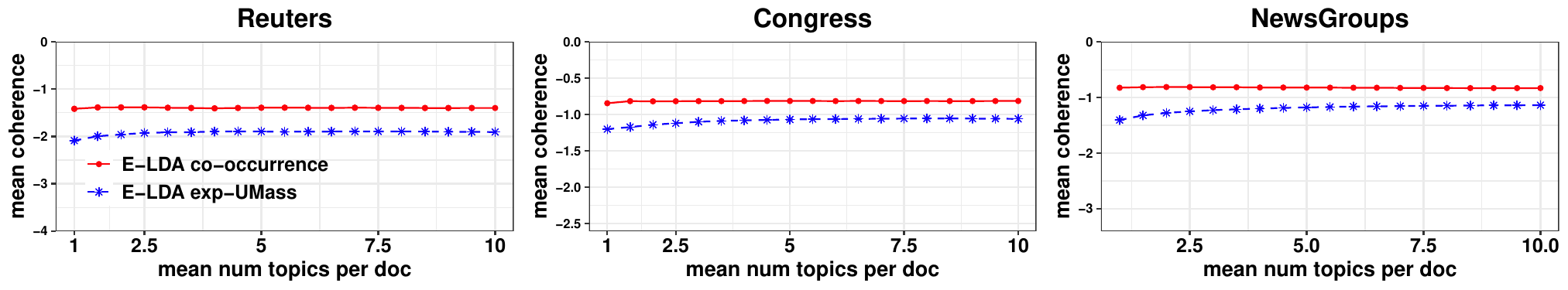}
   \caption[Consistency of E-LDA's coherence]{\textbf{Consistency of E-LDA's coherence} as solutions grow from sparse to dense, $1$$\rightarrow$$10$ topics-per-document.}
   \label{fig:consistencyofcoherence}
\end{subfigure}
% \caption[(Top row) E-LDA vs. additional benchmarks, \& (Bottom row) consistency of E-LDA coherence]{E-LDA vs. additional benchmarks, \& consistency of E-LDA coherence.}
%\label{fig:consistencyofcoherence}
\vspace{-0.5em}
\end{figure*}

\textbf{Experiments Set 2 Results: Topics' Semantic Quality.}
Fig. \ref{fig:manybenchmarks} compares the mean topic coherence of our solutions to baselines. E-LDA \textsc{co-occurrence} and \textsc{exp-UMass} solutions obtain $2.53$-$4.20$ and $1.18$-$1.67$ times better mean coherence, respectively, compared to the Gibbs topics when coherence scores computed based on the $5$ top words per topic (leftmost points in Fig. \ref{fig:manybenchmarks}). This suggests that E-LDA's topics are significantly better than Gibbs topics at conveying a small and closely-related set of `keywords' that are more semantically evocative of their topic's subject matter. As we recompute coherence based on $h^*=10,\dots,25$ top words per topic (rightmost points in Fig. \ref{fig:manybenchmarks}), E-LDA \textsc{co-occurrence} and \textsc{exp-UMass} topics outperform Gibbs by a factor that increases to $2.92$-$3.21$ and $1.93$-$2.33$, respectively. This suggests that each E-LDA topic also captures more subtle semantic relationships beyond a handful of keywords, whereas Gibbs topics tend to exhibit more  noise as we move down the list of their top words. Because coherence is a logarithmic measure, E-LDA's several-fold improvement may be viewed as an exponential improvement in the degree to which E-LDA topics emphasize co-appearing words. Sparse and Bayesian variants of the Gibbs baselines improve slightly over the standard Gibbs on $2$ of the $3$ datasets, but neither obtains the semantic quality of \textsc{co-occurrence} and \textsc{exp-UMass}. Disaggregating the results, we note that (i) E-LDA \textsc{co-occurrence}'s plotted \emph{mean} topic coherence outperforms the \emph{best single topic's} coherence of all Gibbs baselines' topics; and (ii) E-LDA \textsc{co-occurrence} \emph{worst} topic's coherence outperforms the \emph{mean} topic coherence of all Gibbs baselines. This holds regardless of $h^*$.

% \textbf{Comparison with \textsc{BERTopic}.}  E-LDA \textsc{co-occurrence} outperforms \textsc{BERTopic}'s coherence by factors of $3.04$ to $4.70$, and E-LDA \textsc{exp-UMass}  outperforms it by factors of $1.46$ to $2.72$. 

% \textbf{Comparison with \textsc{AVITM}.} E-LDA \textsc{co-occurrence} also outperforms \textsc{AVITM} by factors of $5.96$ to $10.17$, and E-LDA \textsc{exp-UMass} outperforms \textsc{AVITM} by factors of $3.08$ to $7.09$. \textsc{AVITM}'s  poor coherence matches anecdotal experience that this algorithm  obtains good coherence on some datasets but requires customized parameter tuning to obtain competitive performance on others. 

% \textbf{Comparison with \textsc{ETM}}. \textsc{ETM} obtains the best coherence of the baselines. We note that ETM was designed as a coherence-improving algorithm, per Dieng et al. (2020). Nonetheless, E-LDA \textsc{co-occurrence} outperforms ETM by factors of $1.19$ to $2.36$, and E-LDA \textsc{exp-UMass} obtains performance roughly equal to \textsc{ETM} (slightly better on Reuters, slightly worse on NewsGroups). 

 Considering neural and LLM-based baselines, E-LDA \textsc{co-occurrence} outperforms \emph{BERTopic},  \emph{AVITM}, and \emph{FASTopic} by factors of $3.04$-$4.70$, $5.96$-$10.17$, and $2.24$-$5.84$, respectively. It also outperforms \emph{ETM} by factors of $1.19$-$2.36$, despite the fact that ETM was designed as a coherence-improving algorithm. E-LDA \textsc{exp-UMass}  outperforms \emph{BERTopic}, \emph{AVITM}, and \emph{FASTopic} by factors of $1.46$-$2.72$, $3.08$-$7.09$, $1.48$-$2.67$ resp., and compares well to \emph{ETM} (better on Reuters, slightly worse on NewsGroups).

% \textbf{Comparison with \textsc{ETM}}. \textsc{ETM} obtains the best coherence of the baselines. We note that ETM was designed as a coherence-improving algorithm, per Dieng et al. (2020). , and E-LDA \textsc{exp-UMass} obtains performance roughly equal to \textsc{ETM} (slightly better on Reuters, slightly worse on NewsGroups). 

Fig. \ref{fig:manybenchmarks} also plots the coherence of E-LDA initialized with an unmodified \textsc{UMass} topic generator. This basic approach performs relatively poorly (see App. \ref{app:umassbad}), suggesting that E-LDA's high coherence obtained via \textsc{co-occurrence} and \textsc{exp-UMass} topics are indicative of high semantic quality and not simply due to their generating function's similarity to the function used to compute post hoc coherence. We note that the \emph{candidate sets} of all three of our topic generators have identical mean coherence (coherence only depends on  ranking, which is invariant to the exponential). The mean coherence of the candidate sets is -$2.9$, -$1.9$, -$1.8$ for Reuters, Congress, and NewsGroups. Observe that these mean coherences are significantly worse than the mean coherence of \textsc{co-occurrence} and \textsc{exp-UMass} solutions topics in Fig. \ref{fig:manybenchmarks}. This is because our algorithms tend to select high-coherence topics from the candidate set under these (sparser) topic generators.

% %%%%% COHERENCE APPENDIX MANYBENCHMARKS %%%%%%
% \begin{figure}[t]
% \centering
% \includegraphics[width=1.0\textwidth]{coherence_figures/BENCHMARK_plots_APPENDIX_eldak_noGoogle.pdf}
% \caption{}
% %\Description[]{} 
% \label{fig:manybenchmarks}
% \vspace{-0mm}
% \end{figure}     

% \emph{Comparison to additional state-of-the-art benchmarks.} 
% Fig. \ref{fig:manybenchmarks} adds additional benchmarks for comparison, including a Gibbs algorithm with added Bayesian optimization, and Gibbs solver initialized with sparse priors. While these optimizations offer improvements in come cases compared to the standard Gibbs approach, their topic coherences remain significantly worse than E-LDA solutions. 

% %%%%% CONSISTENCY OF COHERENCE %%%%%%
% \begin{figure}[b]
% \centering
% \includegraphics[height=0.3\textwidth]{coherenceonsistency_figures/BENCH_CONSISTENCY_plots_APPENDIX_eldak_noGoogle.pdf}
% \caption{Consistency of E-LDA's coherence scores as we grow its solutions from maximally sparse $1$ topic-per-document to a dense $10$ topics-per-document.}
% %\Description[]{} 
% \label{fig:consistencyofcoherence}
% \vspace{-0mm}
% \end{figure}  
  
\textbf{Consistency of E-LDA's semantic quality advantage across sparse and dense solutions.} 
Fig. \ref{fig:consistencyofcoherence} plots the consistency of E-LDA's topic coherences as we grow its solutions from a maximally sparse arrangement of $1$ topic-per-document (equivalent to a document clustering) to a relatively dense $10$ topics-per-document. Here, we observe that E-LDA's solutions exhibit remarkable consistency in terms of their high coherence regardless of the sparsity of the desired solution size. In short, E-LDA returns solutions of consistently high semantic quality regardless of whether researchers set them to produce a simple and sparse summary of their text data or a nuanced and dense one.

\section{Conclusion}
\label{sec:conclusion}

%\emph{This paper presents work whose goal is to advance the field of Machine Learning. There are many potential societal consequences of our work, none which we feel must be specifically highlighted here.}

In this paper, we showed a novel combinatorial approach to topic modeling that efficiently and near-optimally solves LDA topic-word assignments under explicit sparsity constraints. We also showed that our approach permits practitioners to restrict the learned topics to those that meet standard interpretability criteria. Importantly, these guarantees and interpretability properties also apply when topics are used for causal inference. Finally, we showed that this approach learns topics of high semantic quality.

We have focused on LDA and sparsity constraints because these are the most practical tools for social science research. However, we emphasize that our techniques generalize to many topic model variations. They can also be solved under more sophisticated constraints such as packing or matroid constraints that permit significant flexibility compared to standard algorithms in terms of encoding domain-specific information into text models.

\section*{Acknowledgments}
\label{sec:acknowledgments} The author is grateful for research support from \mbox{OpenAI}, and for valuable comments and suggestions from \mbox{Maximilian} Fox, Kosuke Imai, Alastair Iain Johnston, and Brandon Stewart.

\section*{Impact statement}
\label{sec:impactstatement}
Mainstream social science applications for topic models models include text modeling for ethically fraught areas. For decisionmaking in such areas, it is ethically desirable and, increasingly, legally mandated to adopt models with strong guarantees and/or interpretability properties \citep{madiega2021artificial}. While neural and LLM/BERT-based topic models have recently shown impressive fidelity for text modeling in general, they lack such properties. We see our work as addressing this important gap.

%\newpage

% \section{Acknowledgements}
% This research is supported by NSF Grant 164732. The author is grateful to Maximilian Fox, Kosuke Imai and the Imai Research Group, and Alastair Iain Johnston for insightful comments.

% IMPACT STATEMENT is now in conclusion

%\newpage
%\section{Notes and stuff}
%\singlespacing
%\clearpage
\bibliography{references}
\addcontentsline{toc}{chapter}{References}
\bibliographystyle{plainnat} % fixes numbers in text but no brackets
\appendix
\onecolumn

\section{The LDA Model \& Integrating out $\theta_d$ }
\label{appendix:integratingouttheta}

The LDA model states that the data is a corpus (set) $D$ of documents, where each document $d\in D$ is a multiset of words: $d=[w_1, \dots, w_{|d|}]$. Suppose there are $k$ topics, where each topic is a distribution over all words in the vocabulary. LDA states that the text corpus is generated as follows: For each document $d$, we sample a document length $|d|$ and a document mixture distribution $\theta_d\sim Dirichlet (\alpha)$ over the $k$ topics. Then, to draw each of the $|d|$ words in $d$, we sample a topic $z_{d,i}\in \{1,\dots,k\}$, where $z_{d,i}\sim Categorical(\theta_d)$, and from this topic we sample a word $w_{d,i}\sim Categorical(\phi_{z_{d,i}})$, where $\phi_{z}$ is the distribution over all words corresponding to topic $z$:

\begin{algorithm}[H]
% Core probabilistic model
\caption*{\textbf{LDA generative model}}
\addcontentsline{loa}{algorithm}{Core probabilistic model}

\begin{algorithmic}
     	\INPUT Corpus $D$ of empty documents (multisets) $d_1, \dots, d_{|D|}$, topics $\phi_1\dots\phi_k$, parameters $\alpha, \lambda$\\
     	%\STATE \textbf{for} topic $\tau\in {1, \dots, k}$\\
     	%\STATE \ \ \ \ \textit{draw} topic distribution $\phi_\tau$ over vocabulary V\\ 
    	\STATE \textbf{for} document $d\in D$\\
        \STATE \ \ \ \ \textit{draw} document length $|d|\sim \text{Poisson}(\lambda)$\\
        \STATE \ \ \ \ \textit{draw} document distribution over topics $\theta_d\sim \text{Dirichlet}(\alpha)$\\
    	\STATE \ \ \ \ \textbf{for} document-word index $i=1, \dots, |d|$\\ 
    	\STATE \ \ \ \ \ \ \ \ \textit{draw} topic $z_{d,i} \sim \text{Categorical}(\theta_{d})$\\ 
    	\STATE \ \ \ \ \ \ \ \ \textit{draw} word $w_{d,i} \sim \text{Categorical}(\phi_{z_{d,i}})$\\ 
    	\STATE \ \ \ \ \ \ \ \ $d[i]\gets  w_{d,i}$\\
        \STATE \textbf{return} $D$
  \end{algorithmic}
  \label{alg:CoreProbModel}
\end{algorithm}

Now, denote by $Z$ all latent topics $z_{d,i}$ corresponding to each word in $D$; by $\Theta$ all documents' topic mixtures $\{\theta_1,\dots, \theta_{|D|}\}$; and  by $\Phi$ the set of all topics $\{\phi_1,\dots,\phi_k\}$. The joint model probability is:
\begin{align}
         \label{eq:LDAjoint}
    &\text{P}(D, Z, \Theta, \Phi \ | \alpha, \beta, \lambda) = 
        \prod_{d\in D} \prod^{|d|}_{i=1} Cat (z_{d, i} | \theta_d)  \times
        c_1\prod_{d\in D} Dir(\theta_d|\alpha)  \\
        &\ \ \ \ \ \ \ \ \ \ \ \ \ \ \ \ \ \ \ \ \ \ \ \ \ \ \ \ \ \ \ \ \ \ \ \ \ \ \ \ 
        \times \prod_{d\in D} \prod^{|d|}_{i=1} Cat (w_{d,i} | z_{d,i}, \ \phi_{z_{d,i}})  \times 
       c_2\prod^{k}_{\tau=1}  Dir(\phi_k|\beta_\tau) 
         \times \prod_{d\in D} Poi(|d|\ |\ \lambda) 
         \nonumber % DELETES EQN NUMBER ON 2nd line
         %
         %\frac{\lambda^{|d|} e^{-\lambda}}{ |d|!} 
\end{align}
The second $Dir$ term is the $Dirichlet(\beta)$ prior on topics, and $c_1$ and $c_2$ are normalizing constants.

Documents' topic mixtures $\theta_d$ can be integrated out of LDA's joint probability as follows. We seek:
\begin{align*}
   &\text{P}(Z, \Phi \ | D,  \alpha, \beta)
\propto \int 
\underbrace{
        \prod_{d\in D} \prod^{|d|}_{i=1} Cat (z_{d, i} | \theta_d)  \times
        c_1\prod_{d\in D} Dir(\theta_d|\alpha) } \ d\theta_d \\
        &\ \ \ \ \ \ \ \ \ \ \ \ \ \ \ \ \ \ \ \ \ \ \ \ \ \ \ \ \ \ \ \ \ \ \ \ \ \ \  \ 
        \times
         \prod_{d\in D} \prod^{|d|}_{i=1} Cat (w_{d,i} | z_{d,i}, \ \phi_{z_{d,i}})  \times 
       c_2\prod^{k}_{\tau=1}  Dir(\phi_k|\beta_\tau) 
         \times \prod_{d\in D} Poi(|d|\ |\ \lambda)\\
\end{align*}
Documents are conditionally independent from each other. Denote the last three terms by $\Omega$:
\begin{align*}
&= 
        \prod_{d\in D} \int \prod^{|d|}_{i=1} Cat (z_{d, i} | \theta_d)  \times
        c_1\prod_{d\in D} Dir(\theta_d|\alpha)  \ d\theta_d
        \times
         %\prod_{d\in D} \prod^{|d|}_{i=1} Cat (w_{d,i} | z_{d,i}, \ \phi_{z_{d,i}})  \times 
       \Omega
\end{align*}
Now, denote by $\theta_d,\tau$ and $\alpha_\tau$ the $\tau$'th elements of $\theta_d$ and $\alpha$, respectively. Substituting the standard Dirichlet prior, we have:
\begin{align*}
&=\prod_{d\in D} \int 
\frac{\Gamma(\sum^k_{\tau=1} \alpha_\tau)}{\prod^k_{\tau=1}\Gamma(\alpha_\tau )}
\prod^k_{\tau=1} \theta_{d,\tau}^{a_\tau - 1} \prod^{|d|}_{i=1} \theta_{d,z_d} \ d \theta_d
        \times
         %\prod_{d\in D} \prod^{|d|}_{i=1} Cat (w_{d,i} | z_{d,i}, \ \phi_{z_{d,i}})  \times 
       \Omega
\end{align*} Changing variables in the product over each documents' words, and letting  $n_{d,\tau}$ denote the count of words in document $d$ that are assigned to topic $\tau$, so $\forall d,t,$ we have $0\le n_{d,\tau} \le |d|; \ \ \forall |d|, \sum^k_{\tau=1} n_{d,\tau} = |d|$. We have:
\begin{align*}
&=\prod_{d\in D} \int \frac{\Gamma(\sum^k_{\tau=1} \alpha_\tau)}{\prod^k_{\tau=1}\Gamma(\alpha_\tau )} \prod^k_{\tau=1} \theta_{d,\tau}^{a_\tau - 1} \prod^{k}_{\tau=1} \theta^{n_{d,\tau}}_{d,\tau} \ d \theta_d
        \times
         %\prod_{d\in D} \prod^{|d|}_{i=1} Cat (w_{d,i} | z_{d,i}, \ \phi_{z_{d,i}})  \times 
       \Omega\\
&=\prod_{d\in D} \int \frac{\Gamma(\sum^k_{\tau=1} \alpha_\tau)}{\prod^k_{\tau=1}\Gamma(\alpha_\tau )} \prod^k_{\tau=1} \theta_{d,\tau}^{a_\tau +n_{d,\tau} - 1} \ d \theta_d
% %
        \times
         %\prod_{d\in D} \prod^{|d|}_{i=1} Cat (w_{d,i} | z_{d,i}, \ \phi_{z_{d,i}})  \times 
      \Omega\\
\end{align*}Multiplying by a fraction that is equal to $1$:
\begin{align*}
&=\prod_{d\in D} 
\frac{\Gamma(\sum^k_{\tau=1} \alpha_\tau)}{\prod^k_{\tau=1}\Gamma(\alpha_\tau )} 
\frac{\prod^k_{\tau=1}\Gamma(n_{d,\tau} + \alpha_\tau )}{\Gamma(\sum^k_{\tau=1} n_{d,\tau} + \alpha_\tau)}
\int \frac{\Gamma(\sum^k_{\tau=1} n_{d,\tau} + \alpha_\tau)}{\prod^k_{\tau=1}\Gamma(n_{d,\tau} + \alpha_\tau )}
\prod^k_{\tau=1} \theta_{d,\tau}^{a_\tau +n_{d,\tau} - 1} \ d \theta_d
%\times \prod_{d\in D} \prod^{|d|}_{i=1} Cat (w_{d,i} | z_{d,i}, \ \phi_{z_{d,i}})  \times 
      %c_2\prod^{k}_{\tau=1}  Dir(\phi_k|\beta_\tau) \times 
      \times \Omega
\end{align*}
Observe that the integral is now over the entire domain of a Dirichlet PDF, so it evaluates to $1$:
\begin{align}
&=\prod_{d\in D} 
\frac{\Gamma(\sum^k_{\tau=1} \alpha_\tau)}{\prod^k_{\tau=1}\Gamma(\alpha_\tau )} 
\frac{\prod^k_{\tau=1}\Gamma(n_{d,\tau} + \alpha_\tau )}{\Gamma(\sum^k_{\tau=1} n_{d,\tau} + \alpha_\tau)} \notag\\
&\ \ \ \ \ \ \ \ \ \ \ \ \ \ \ \ \ \ \ \ \ \ \ \ 
        \times
         \prod_{d\in D} \prod^{|d|}_{i=1} Cat (w_{d,i} | z_{d,i}, \ \phi_{z_{d,i}})  \times 
      c_2\prod^{k}_{\tau=1}  Dir(\phi_k|\beta_\tau) 
         \times \prod_{d\in D} Poi(|d|\ |\ \lambda)
         \label{eq:LDAmarginal}
\end{align}

Ignoring the nuisance document length term (which is standard \citep{blei2003latent}), then for a fixed set of topics $\Phi$ we arrive at the canonical MAP topic assignment inference problem studied in seminal work by \citet{sontag2011complexity}. Taking logarithms of the following gives the expression in Section \ref{sec:LDAbackground}:

\begin{align}
    \text{P}(Z | D) \propto
        \displaystyle \prod_{d \in D} \Big(
        \frac{\Gamma(\sum^k_{\tau=1} \alpha_\tau)\prod^k_{\tau=1}\Gamma(n_{d,\tau} + \alpha_\tau)}{\Gamma(\sum^k_{\tau=1}\alpha_\tau + |d|) \prod^k_{\tau=1} \Gamma(\alpha_\tau)}
          \prod^{|d|}_{i=1} Cat (w_{d,i} | z_{d,i}, \ \phi_{z_{d,i}})   \Big)
        %\frac{\lambda^{|d|} e^{-\lambda}}{ |d|!}
        \label{appendixeq:MAP}
\end{align}

\section{Deferred proof of Theorem \ref{theorem:infinitealpha}}
\label{appendix:InfiniteLimitProof}

\begin{theorem*}
\begin{align}
\displaystyle \lim_{\alpha\rightarrow\infty} \big( \underset{Z}{\mathrm{argmax}} \ \log \text{P}(Z|D) \big) = \underset{Z}{\mathrm{argmax}} \sum_{d \in D} \sum^{|d|}_{i=1} \log Cat (w_{i,d} | z_{i,d}, \ \phi_{z_{i,d}})
\end{align}
\end{theorem*}

\emph{Proof.} From Section \ref{eq:logMAP} we have:
\begin{align}
% \log \text{P}(Z|D)\propto
%         \sum_{d \in D}\Big(\sum^k_{\tau=1} \log \Gamma(n_{d,\tau} + \alpha_\tau) +
%         \log \Gamma(\sum^k_{\tau=1} \alpha_\tau) 
%          - \log \Gamma(\sum^k_{\tau=1} \alpha_\tau + |d|) - \sum^k_{\tau=1} \log \Gamma(\alpha_\tau)\\ + \sum^{|d|}_{i=1} \log Cat (w_{d,i} | z_{d,i},  \phi_{z_{d,i}})  \Big)
%         %\label{eq:logMAP} it's labeled in its first instance!
% \end{align}
% \begin{align}
 \log \text{P}(Z|D)\propto  \sum_{d \in D} \Big( &\underbrace{\sum^{|d|}_{i=1} \log Cat (w_{d,i} | z_{d,i},  \phi_{z_{d,i}})}_{likelihood} + \underbrace{\sum^k_{\tau=1} \log \Gamma(n_{d,\tau} + \alpha_\tau) }_{sparsity}\notag \\[-4pt]  &\enspace+\underbrace{\log \Gamma(\sum^k_{\tau=1} \alpha_\tau) - \log \Gamma(\sum^k_{\tau=1} \alpha_\tau + |d|) - \sum^k_{\tau=1} \log \Gamma(\alpha_\tau)}_{constant}
            \Big)
         \label{eq:logMAPappendix}
\end{align}

We will use the following well-known asymptotic property of $\Gamma$ functions: for $r\in\mathbb{C}$, $lim_{r\rightarrow \infty}\Gamma(q+r) = \Gamma(q)q^r$. After taking logs and rearranging, we have the property $lim_{r\rightarrow \infty}\left[\log \frac{\Gamma(kq+r)}{ \log (\Gamma(kq))} - r\log(kq)\right] = 0$ for $k>0$.

\newpage % otherwise breaks up the explanation

Consider the limit for a single document, and let $\mathcal{L}$ denote the likelihood term that is independent of $\alpha$. Rearranging, we have:

\begin{align*}
&\lim_{\alpha\rightarrow\infty} \Big[- \log \frac{\Gamma(\sum_{\tau=1}^k\alpha_i + |d|)}{\Gamma(\sum_{\tau=1}^k\alpha_i)}  + \sum^k_{\tau=1} \log\frac{\Gamma(\alpha_i+n_{d,\tau} )}{\Gamma(\alpha_i)} + \mathcal{L}\Big]\\
=&\lim_{\alpha\rightarrow\infty} \Big[- \log \frac{\Gamma(k\alpha + |d|)}{\Gamma(k\alpha)}  + \sum^k_{\tau=1} \log\frac{\Gamma(\alpha+n_{d,\tau} )}{\Gamma(\alpha)} + 0 - \sum^k_{\tau=1} 0\Big] + \mathcal{L}\\
&\text{Applying our asymptotic property with $q=k\alpha,\ r=|d|$ and noting $k>0$ we have:}\\
=&\lim_{\alpha\rightarrow\infty} \Big[ - \log \frac{\Gamma(k\alpha + |d|)}{\Gamma(k\alpha)}  + \sum^k_{\tau=1} \log\frac{\Gamma(\alpha+n_{d,\tau} )}{\Gamma(\alpha)} \\ & \ \ \ \ \ \ \ \ \ \ \ \ \ \ \ \ \ \ \ \ \ + \Big(\log\frac{\Gamma(k\alpha+|d|)}{\Gamma(k\alpha)}-|d|\log (k\alpha) \Big) - \sum^k_{\tau=1} 0 \Big] + \mathcal{L}\\
%
%&\lim_{\alpha\rightarrow\infty} \Big[  \sum^k_{\tau=1} \log\frac{\Gamma(\alpha_i+n_{d,\tau} )}{\Gamma(\alpha_i)} + -|d|\log (k\alpha)  + \sum^k_{\tau=1} 0\Big]\\
%
=&\lim_{\alpha\rightarrow\infty} \Big[  \sum^k_{\tau=1} \log\frac{\Gamma(\alpha+n_{d,\tau} )}{\Gamma(\alpha)}  -|d|\log (k\alpha)  - \sum^k_{\tau=1} 0\Big] + \mathcal{L}\\
&\text{Applying our asymptotic property $k$ times with $q=\alpha, \ r=n_{d,\tau}$ for $\tau=1\dots k$  we have:}\\
=&\lim_{\alpha\rightarrow\infty} \Big[  \sum^k_{\tau=1} \log\frac{\Gamma(\alpha+n_{d,\tau} )}{\Gamma(\alpha)}  -|d|\log (k\alpha)   -   \Big(\sum_{\tau=1}^k \Big(\log\frac{\Gamma(\alpha+n_{d,\tau})}{\Gamma(\alpha)}-n_{d,\tau}\log \alpha \Big) \Big) \Big] + \mathcal{L}\\
%
%&\lim_{\alpha\rightarrow\infty} \Big[  \sum^k_{\tau=1} \log\frac{\Gamma(\alpha_i+n_{d,\tau} )}{\Gamma(\alpha_i)} + -|d|\log (k\alpha)  +   \Big(\sum_{\tau=1}^k \log\frac{\Gamma(\alpha_i+n_{d,\tau})}{\Gamma(\alpha_i)}-\sum_{\tau=1}^k n_{d,\tau}\log \alpha_i \Big) \Big]\\
%
%=&\lim_{\alpha\rightarrow\infty} \Big[  \sum^k_{\tau=1} \log\frac{\Gamma(\alpha+n_{d,\tau} )}{\Gamma(\alpha)}  -|d|\log (k\alpha)   +   \Big(-\sum_{\tau=1}^k \log\frac{\Gamma(\alpha+n_{d,\tau})}{\Gamma(\alpha)}+\sum_{\tau=1}^k n_{d,\tau}\log \alpha \Big)  \Big] + \mathcal{L}\\
%
=&\lim_{\alpha\rightarrow\infty} \Big[   -|d|\log (k\alpha)   +   \sum_{\tau=1}^k n_{d,\tau}\log \alpha   \Big] + \mathcal{L}\\
=&\lim_{\alpha\rightarrow\infty} \Big[   -|d|\log (k\alpha)   +   |d|\log \alpha   \Big]  + \mathcal{L}\\
=&\lim_{\alpha\rightarrow\infty} \Big[   -|d|\log k      \Big] + \mathcal{L}\\
= & -|d|\log k + \mathcal{L}\\
&\text{After substituting for $\mathcal{L}$ and adding up this single-document limit over all documents we have:}\\
&  \displaystyle \lim_{\alpha\rightarrow\infty} \big( \underset{Z}{\mathrm{argmax}} \ \log \text{P}(Z|D) \big) =   \underset{Z}{\mathrm{argmax}} \Big[ -\log(k) \cdot \displaystyle \sum_{d\in D}|d| + \displaystyle  \sum_{d\in D} \displaystyle  \sum^{|d|}_{i=1} \log Cat (w_{i,d} | z_{i,d}, \ \phi_{z_{i,d}}) \Big]\\
& = \underset{Z}{\mathrm{argmax}}  \displaystyle  \sum_{d\in D} \displaystyle  \sum^{|d|}_{i=1} \log Cat (w_{i,d} | z_{i,d}, \ \phi_{z_{i,d}}) \text{ \ \ as the left term above is const. independent of $\alpha$.}
\end{align*}

\section{Deferred proof of submodularity}
\label{AppendixSubmodular}

Recall that original objective function $f$ from eq. \ref{eq:ETM} is:

\begin{align}
\text{\emph{Objective function:} } f(E) &= \max \big[ \log P(Z|D,E) \big]  \\ & = \sum_{d\in D}\sum_{i=1}^{|d|} \ \max_{\tau\in E_d} (\log \phi_\tau[w_{d,i}])\notag\\[-0.1em]
\text{find } \underset{{\{\tau\Rightarrow d\}, \tau \in \Phi, d\in D}}{\argmax} f(E) &\text{\ s.t.}  \ |E|\le \kappa|D|; \  |E_d|\ge 1, \forall d \notag
\end{align}

We noted that one challenge is that $f$ is undefined when some document has no links in $E$. We addressed this by extending the initialization technique of \citet{gomes2010budgeted} by linking each document to a single improper placeholder topic $p$ with infinitesimal probability on each word, then maximizing $\dot{f}$ (equivalent to maximizing $f$): $\dot{f}(E) = f(E \cup P) - f(P)$, where $P$ is the set of placeholder topics.

We now show that objective $\dot{f}(E)$ is a monotone submodular function. First, observe that the objective $\dot{f}$ is \emph{monotone}, as adding more links to $E$ cannot decrease $\dot{f}(E)$ due to the $\max$ function.
 Now, consider the \emph{\textbf{marginal value}} $\dot{f}_{E}(A) \coloneqq \dot{f}(E \cup A) - \dot{f}(E)$ of adding links $A = \{[\tau \Rightarrow d]\}$ to $E$.

\vspace{-2em}
\begin{align}
\dot{f}_{E}(A) \coloneqq& \dot{f}(E \cup A) - \dot{f}(E) \ \ \text{\ \ \ \ \ is our marginal value function}. \notag\\
=& f(E \cup A \cup P) - f(P) - \big( f(E \cup P) - f(P) \big)\notag \\
=& \sum_{d\in D}\sum_{i=1}^{|d|} \ \max_{\tau^{'}\in (E_d \cup A \cup P)} (\log \phi_{\tau^{'}}[w_{d,i}]) - \sum_{d\in D}\sum_{i=1}^{|d|} \ \max_{\tau^{''}\in P} (\log \phi_{\tau^{''}}[w_{d,i}])  \notag\\ 
& \ \ \ \ \ \ \ - \Big( \sum_{d\in D}\sum_{i=1}^{|d|} \ \max_{\tau\in (E_d \cup P)} (\log \phi_\tau[w_{d,i}]) - \sum_{d\in D}\sum_{i=1}^{|d|} \ \max_{\tau^{''}\in P} (\log \phi_{\tau^{''}}[w_{d,i}]) \Big)\notag\\
=& \sum_{d\in D}\sum_{i=1}^{|d|} \ \max_{\tau^{'}\in (E_d \cup A \cup P)} (\log \phi_{\tau^{'}}[w_{d,i}])  - \sum_{d\in D}\sum_{i=1}^{|d|} \ \max_{\tau\in (E_d \cup P)} (\log \phi_\tau[w_{d,i}]) \label{multidoc}\\
&\text{For a single document $d$, we have:}\notag\\
=& \sum_{i=1}^{|d|} \ \max_{\tau^{'}\in (E_d \cup A \cup P)} (\log \phi_{\tau^{'}}[w_{d,i}])  - \sum_{i=1}^{|d|} \ \max_{\tau\in (E_d \cup P)} (\log \phi_\tau[w_{d,i}])\notag\\
%
% &\text{As before, denote $z_{d,i} = \argmax_{\tau \in T_d} \phi_\tau[w_{d,i}]$}\\
% %
% =& \sum_{i=1}^{|d|} \ \max_{\tau\in (E_d \cup A \cup P)} (\log \phi_\tau[w_{d,i}])  - \sum_{i=1}^{|d|} \ \max_{\tau\in (E_d \cup P)} (\log \phi_\tau[w_{d,i}])\\
%
=& \sum_{i=1}^{|d|} \Big( \ \max_{\tau^{'}\in (E_d \cup A \cup P)} (\log \phi_{\tau^{'}}[w_{d,i}])  -   \max_{\tau\in (E_d \cup P)} (\log \phi_\tau[w_{d,i}]) \Big)\notag \\
%
% =& \sum_{i=1}^{|d|} \Big( \big( \ \max_{\tau^{'}\in A} (\log \phi_{\tau^{'}}[w_{d,i}])  -   \max_{\tau\in (E_d \cup P)} (\log \phi_\tau[w_{d,i}]) \big) \mathbf{1}\big (\max_{\tau^{'}\in A} (\log \phi_{\tau^{'}}[w_{d,i}]) >  \max_{\tau\in (E_d \cup P)} (\log \phi_\tau[w_{d,i}])\big) \Big) \\
%
&\text{Denote by $\mathbf{1}(\cdot)$ a function that takes the value of $1$ if its argument is true and $0$ otherwise. Now:}\notag\\
=&\sum_{i=1}^{|d|} \Big( \big( \ \max_{\tau^{'}\in A} (\log \phi_{\tau^{'}}[w_{d,i}])  -   \max_{\tau\in (E_d \cup P)} (\log \phi_\tau[w_{d,i}]) \big) \cdot \mathbf{1}\big (\max_{\tau^{'}\in A} (\phi_{\tau^{'}}[w_{d,i}]) >  \max_{\tau\in (E_d \cup P)} ( \phi_\tau[w_{d,i}])\big) \Big) 
\end{align}

Observe that as the set $E_d$ grows (i.e. $E_d$ is expanded to include more topics linked to $d$), the first term, $\max_{\tau^{'}\in A} (\log \phi_{\tau^{'}}[w_{d,i}])$ is fixed and the second term, $\max_{\tau\in (E_d \cup P)} (\log \phi_\tau[w_{d,i}])$, is nondecreasing. Therefore, observe (1) the sum $\sum_{i=1}^{|d|}  \big( \ \max_{\tau^{'}\in A} (\log \phi_{\tau^{'}}[w_{d,i}])  -   \max_{\tau\in (E_d \cup P)} (\log \phi_\tau[w_{d,i}]) \big)$ is nonincreasing.  Then observe (2) that any topic $\tau\in A$ will be the maximum-probability topic for (weakly) fewer words in $d$, thus it will contribute (weakly) fewer terms to $\dot{f}$ due to the $\mathbf{1}(\cdot)$ function. Putting these together, we have that the marginal value of adding $A$ to $E_d$ is weakly decreasing as $E_d$ grows:  $\dot{f}_{E_d}(A) \geq \dot{f}_{E'_d}(A), \ \  \forall E_d \subseteq E'_d, \ \tau \in A \subseteq \Phi, \text{ and } d\in D$.

Finally, this analysis supposed a corpus of a single document. However, per line \ref{multidoc}, $\dot{f}$ is a sum of $|D|$ per-document submodular functions. Sums of submodular functions are also submodular.

\vspace{5em}

\section{Unconstrained problem formulations \& expected number of unique topics per document}
\label{appendix:expectednotopics}

  Observe that in the unconstrained case (i.e. where each document can have unlimited topics up to one unique topic per word), then finding $d$'s MAP topic-word assignments, $\argmax_{Z_d} \text{P}(Z_d|d)$, is easy. In that case, we just assign each word in the data to the topic in $\Phi$ with the highest probability of drawing it: $z_{d,i} = \argmax_{\tau \in \Phi} \phi_\tau[w_{d,i}]$. Summing these, we have: 
\begin{align}
 \max_{Z} \big[ \log P(Z|D)\big] = \sum_{d\in D}\sum_{i=1}^{|d|} \ \max_{\tau\in \Phi} (\log \phi_\tau[w_{d,i}])   
%\label{eq:easy}
\end{align}

However, this unconstrained problem returns practically unusable, maximally overfit solutions where each word in a document may be associated with a different topic. In the analysis below, we show below that in the unconstrained problem, the expected number of unique topics-per-document is well-approximated by $|\Phi|(1 - e^{-|d|/|\Phi|})$. For practical-sized datasets such as (for example) Twitter datasets commonly used in social science research, $|\Phi|$ exceeds the word count of any single document (Tweet). In that case, the unconstrained MAP solution assigns each document a number of unique topics that approaches one unique topic-per-word. In other words, the \emph{unconstrained} MAP solution returns a `worst-case overfit'. Even when the number of topics is the same as the wordcount of a document (e.g. $100$), the unconstrained problem's MAP solution assigns each $100$-word tweet $|\Phi|(1 - e^{-|d|/|\Phi|}) = 100(1 - e^{-100/100}) \approx 63.2$ unique topics per $100$-word document in expectation. As such, the unconstrained problem formulation yields unusably dense solutions for practical problem instances. 

Indeed, as we discussed in Section \ref{sec:LDAbackground}, one of the primary problems with using documents' underlying distributions $\theta_d$ for downstream social science analysis is that for many practical datasets such as Twitter data, the number of topics required to adequately model the dataset exceeds the length of any document: $k>|d|$. As such, a document's underlying topic distribution vector $\theta_d$ is not a dimensionality reduction (which is often the goal of using LDA in social science application domains), but a dimensionality expansion. Absent a sparsity constraint, the MAP topic-word assignments exhibit similarly undesirable dimensionality. These issues motivate our sparsity-constrained problem formulation.

More formally, we obtain the expected average number of topics per document in the MAP solution for a given \emph{unconstrained} problem (i.e. where each document can have unlimited topics up to one unique topic per word) as follows. For a single document, in the infinite limit of $\alpha$ the probability of drawing zero words from topic $\tau$ is $\Big(\frac{(|\Phi|-1)} {|\Phi|}\Big)^{|d|}$. The probability of drawing at least one word from topic $\tau$ is the complement $1-\Big(\frac{(|\Phi|-1)} {|\Phi|}\Big)^{|d|}$. Therefore the expected number of unique topics in document $d$ is $|\Phi| \Big[ 1-\Big(\frac{(|\Phi|-1)} {|\Phi|}\Big)^{|d|} \Big]$. To understand how this scales, we can rearrange and use the standard limit approximation to obtain the useful approximation $|\Phi| \Big[ 1-\Big(\frac{(|\Phi|-1)} {|\Phi|}\Big)^{|d|} \Big] \approx |\Phi|(1 - e^{-|d|/|\Phi|})$.

To obtain the expected average number of unique topics per document in the corpus, we just take the expectation over all documents, $\frac{1}{|D|}\sum_{d\in D} |\Phi| \Big[ 1-\Big(\frac{(|\Phi|-1)} {|\Phi|}\Big)^{|d|} \Big] = 
%\frac{|\Phi|}{|D|}\sum_{d\in D}  \Big[ 1-\Big(\frac{(|\Phi|-1)} {|\Phi|}\Big)^{|d|} \Big] = 
|\Phi|(1-1/|D|)\sum_{d\in D} (1-1/|\Phi|)^{|d|}$. Applying this to our approximation gives $ |\Phi|(1-1/|D|)\sum_{d\in D} (1-1/|\Phi|)^{|d|}   \approx 
 \frac{\Phi}{|D|}\sum_{d\in D}(1 - e^{-|d|/|\Phi|})
 $.
  % $\Phi/|D|\sum_{d\in D}(1 - e^{-|d|/|\Phi|})$

\medskip
\textbf{Unconstrained problem with known $\theta$.} Alternatively one could also consider a variant of the unconstrained problem formulation where we assume that documents' underlying topic distributions $\theta_d$ are also known. In that case, we could easily obtain the MAP topic-word assignments by maximizing:
\begin{align}
 \max_{Z} \big[ \log P(Z|D)\big] = \sum_{d\in D}\sum_{i=1}^{|d|} \ \max_{\tau\in \Phi} (\log \theta_d[\tau]\cdot\phi_\tau[w_{d,i}])  
\end{align}
However, the assumption that $\theta_d$ are known is very strong. Mainstream algorithms such as Gibbs provide no guarantees for $\theta_d$. It is possible to obtain estimates of $\theta_d$ via e.g. the algorithms of \citet{arora2016provable}, but these algorithms yield $\theta_d$ with errors that are $o(1/|E_d|)$. In other words, for that approach, solutions that are more desirably sparse exhibit \emph{more} errors in terms of $\theta_d$, yielding more errors in terms of the resulting MAP topic-word assignments. We contrast these problems with the near optimal deterministic guarantees of our approach.

\section{Simple Greedy Algorithm}
\label{app:approximationguarantee}

Consider the simple Greedy algorithm that grows the solution set of topic-to-document links $E$ by iteratively adding the link with the greatest marginal contribution to $\dot{f}$. We formalize this algorithm and derive its complexity below. While it is practically infeasible, it has near-optimal global and per-document approximation guarantees. Our optimized \textsc{FastGreedy-E-LDA} algorithm accelerates this simple Greedy algorithm while maintaining the same guarantees.

\begin{algorithm}[H]
%\caption{2-Class Directed Preferential Attachment}
\caption*{\textbf{Simple-Greedy-E-LDA}}
\addcontentsline{loa}{algorithm}{Simple-Greedy-E-LDA}

% \begin{algorithmic}
%     	\INPUT  Corpus $D$, Candidate set of topics $\Phi$, Sparsity constraint upper-bound $\kappa$ \\
%     	\STATE \textbf{initialize} $E \gets \{[\tau_{phantom}\Rightarrow 1], [\tau_{phantom}\Rightarrow 2],\dots\, [\tau_{phantom}\Rightarrow |d|]\}$ %topic-word assignments  $Z_{MAP}:z_{d,i} \gets \phi_{phantom}, \forall [d,i]$\\
%     	\STATE \textbf{for} $t \in \{1,\dots,\kappa|D|\}$ steps \textbf{do}\\
%         %\STATE \ \ \ \ $X \gets$ a random subset of candidate topics $\phi$ in $\Phi\backslash \Phi_{MAP}$
%         \STATE \ \ \ \ \ $[\tau^*\Rightarrow d^*] \gets$ the topic-doc link that adds the highest marginal value,  $\underset{\tau \in \Phi, d\in D}\argmax \  f_{E}([\tau\Rightarrow d])$\\
%         %\STATE $Z \gets \argmax \ \big[ \sum_{d\in D}\sum_{i=1}^{|d|} \ \max_{\tau\in E_d} (\log \phi_\tau[w_{d,i}]) \big]$\\
%         \STATE \ \ \ \ \ $E \gets [\tau^*\Rightarrow d^*]$
%         \STATE \textbf{return} \ \  $E \backslash \{[\tau_{phantom}\Rightarrow 1], [\tau_{phantom}\Rightarrow 2],\dots\, [\tau_{phantom}\Rightarrow |d|]\}$%,\ \  $Z$ 
%   \end{algorithmic}
%   \label{alg:Greedy}
% \end{algorithm}
% %$\Phi_R$ stands for realized; X is the random set; 
% Where $\phi_{phantom}$ is an phantom topic with constant (improper) small probability on all words, e.g. $[1^{-10}]^{|V|}$.

\begin{algorithmic}
    	\INPUT  Corpus $D$, Candidate topic set $\Phi$, Sparsity constraint upper-bound $\kappa$, Phantom topic links $\mathcal{P}$ \\
    	\STATE \textbf{initialize} $E \gets \mathcal{P}$ %topic-word assignments  $Z_{MAP}:z_{d,i} \gets \phi_{phantom}, \forall [d,i]$\\
    	\STATE \textbf{for} $\kappa|D|$ steps \textbf{do}\\
        %\STATE \ \ \ \ $X \gets$ a random subset of candidate topics $\phi$ in $\Phi\backslash \Phi_{MAP}$
        \STATE \ \ \ \ \ $[\tau^*\Rightarrow d^*] \gets$ the topic-doc link that adds the highest marginal value,  $\underset{\tau \in \Phi, d\in D}\argmax \  f_{E}([\tau\Rightarrow d])$\\
        %\STATE $Z \gets \argmax \ \big[ \sum_{d\in D}\sum_{i=1}^{|d|} \ \max_{\tau\in E_d} (\log \phi_\tau[w_{d,i}]) \big]$\\
        \STATE \ \ \ \ \ $E \gets [\tau^*\Rightarrow d^*]$
        \STATE \textbf{return} \ \  $E \backslash \mathcal{P}$%,\ \  $Z$ 
  \end{algorithmic}
  \label{alg:Greedy}
\end{algorithm}
%$\Phi_R$ stands for realized; X is the random set; 
Where each phantom topic $\phi_{phantom}$ in the phantom topic links $\mathcal{P}$ is a topic with constant (improper) small probability on all words, e.g. $[1^{-10}]^{|V|}$.

\paragraph{\textbf{1-1/e Approximation Guarantee.}}
For completeness, we show \textsc{Simple-Greedy}'s $1-1/e$ approximation guarantee via the classic result of \citet{nemhauser1978best}. 
%
%
%  Let $E^{OPT}$ denote the optimal (maximum value) solution of size $\kappa|D|$, with value $\dot{f}(E^*)=OPT$. We claim via induction that for $0\le i \le \kappa|D|$:
%
% \begin{align}
%     f(O) - f(S_i) \le (1-1/k)^if(O)  \label{hypoth}
% \end{align}
%
% This is trivially true for $i=0$. First, consider \textsc{Simple-Greedy}'s solution $S_{i-1}$ at the last iteration $i-1$. Because there are $|O\backslash S_{i-1}| \le k$ more elements in $O$ than in $S_{i-1}$, and the `best' one we can add to $S_{i-1}$ now is what \textsc{Simple-Greedy} will add first, then we have (by subadditivity):
% \begin{align}
%     f(O) - f(S_{i-1}) \le \sum_{a \in O \backslash S_{i-1}} f_{S_{i-1}}(a) \notag \\
% \frac{1}{k}\big(f(O) - f(S_{i-1})\big) \le f_{S_{i-1}}(a_i) \label{useful}
% \end{align}
%
% Now, returning to our hypothesis, we can expand left-hand-side (i.e. `separate out' the last link that was added to \textsc{Simple-Greedy}'s solution at iteration $i$):
% \begin{align} 
% f(O) - f(S_i)  &= f(O) - f(S_{i-1}) - f_{S_{i-1}}(a_i) \notag\\ 
% &\le f(O) - f(S_{i-1}) - \frac{1}{k}\big(f(O) - f(S_{i-1})\big) \text{ \  substituting eq. \ref{useful}} \notag\\
% &=(1-1/k)(f(O) - f(S_{i-1})\notag\\
% & \le (1-1/k)^i f(O)\notag
% \end{align}
%
%
% \noindent Now set $i=k$, rearrange, and use $(1-1/k)^k \le 1/e$ to obtain $f(S) \ge (1-1/e)f(O)$
%
%
Let $E^{OPT}$ denote the optimal (maximum value) solution of size $\kappa|D|$, with value $\dot{f}(E^{OPT})=OPT$. Let $E^{i}$ denote \textsc{Simple-Greedy}'s solution at iteration $i$. We claim via induction that for its iterations $0\le i \le \kappa|D|$:
\begin{align}
    \dot{f}(E^{OPT}) - \dot{f}(E^i) \le \Big(1-\frac{1}{\kappa|D|}\Big)^i \dot{f}(E^{OPT})  \label{hypoth}
\end{align}
This is trivially true for $i=0$. First, consider \textsc{Simple-Greedy}'s solution $E^{i-1}$ at a previous iteration $i-1$. Because there are $|O\backslash E^{i-1}| \le \kappa|D|$ more elements in $E^{OPT}$ than in $E^{i-1}$, and the `best' one we can add to $E^{i-1}$ now is what \textsc{Simple-Greedy} will add first, then we have (by subadditivity):
\begin{align}
    \dot{f}(E^{OPT}) - \dot{f}(E^{i-1}) \le \sum_{a \in E^{OPT} \backslash E^{i-1}} \dot{f}_{E^{i-1}}(a) \notag \\
\frac{1}{\kappa|D|}\big(\dot{f}(E^{OPT}) - \dot{f}(E^{i-1})\big) \le \dot{f}_{E^{i-1}}(a_i) \label{useful}
\end{align}
Now, returning to our hypothesis, we can expand left-hand-side (i.e. `separate out' the last link that was added to \textsc{Simple-Greedy}'s solution at iteration $i$):
\begin{align} 
\dot{f}(E^{OPT}) - \dot{f}(E^{i})  &= \dot{f}(E^{OPT}) - \dot{f}(E^{i-1}) - \dot{f}_{E^{i-1}}(a_i) \notag\\ 
&\le \dot{f}(E^{OPT}) - \dot{f}(E^{i-1}) - \frac{1}{\kappa|D|}\big(\dot{f}(E^{OPT}) - \dot{f}(E^{i-1})\big) \text{ \  substituting eq. \ref{useful}} \notag\\
&=\big(1-\frac{1}{\kappa|D|}\big)\big(\dot{f}(E^{OPT}) - \dot{f}(E^{i-1})\big)\notag\\
& \le \big(1-\frac{1}{\kappa|D|}\big)^i \dot{f}(E^{OPT})\notag
\end{align}
\noindent Now set $i=\kappa|D|$, rearrange, and use $(1-\frac{1}{\kappa|D|})^{\kappa|D|} \le 1/e$ to obtain $\dot{f}(E) \ge (1-1/e)OPT$.

% \paragraph{\textbf{Complexity of \textsc{Simple-Greedy}.}}
\section{Complexity of \textsc{Simple-Greedy}}
\label{app:simplegreedy}

We now derive the complexity of the vanilla \textsc{Simple-Greedy} algorithm described in Section \ref{sec:submodular} and  compare it to the \textsc{Accelerated-Greedy-E-LDA} algorithm described here:

Simple-Greedy completes in $\kappa|D|$ iterations. Each iteration $i$ of Simple-Greedy computes the marginal value of $|\Phi||D| - i$ links. Without the speedups described in \textsc{FastGreedy-E-LDA}, computing one of these marginal values entails solving $f(E^i \cup [\tau\Rightarrow d]) - f(E^i)$, where $[\tau\Rightarrow d]$ is a new topic-doc link (where $E^i$ denotes the solution at iteration $i$). Computing a value of $f$ requires, for each document in $D$, finding the max log probability of each word in $d$ among $E_d$ topics (or $E_d+1$ topics for $f(E^i \cup [\tau\Rightarrow d])$), then summing the result. Denote the number of words in the longest document by $\ell$. Thus each marginal value requires $O(\ell|E_d||D|)$ which is $O(\ell\kappa|D|)$. An iteration of Simple-Greedy requires computing $O(|\Phi||D|)$ such marginal values, so each iteration is $O(\ell\kappa|\Phi||D|^2)$.

\section{\textsc{FastGreedy-E-LDA} details and analysis}
\label{app:fastgreedy}

We now describe the main ideas we use to obtain \textsc{FastGreedy-E-LDA}'s fast theoretic and practical runtimes while maintaining the deterministic near-optimal guarantees of the vanilla \textsc{SimpleGreedy} algorithm. We analyze complexity below.

\begin{itemize}
    \item \textbf{Improving the per-iteration complexity by the first $|D|$-factor.}  Simple-Greedy evaluates the marginal value of all $|D||\Phi|$ links each iteration. These costly operations can be avoided through thoughtful design. Specifically, after a link $[\tau^*\Rightarrow d^*]$ is added to the solution set $E$, the topic-word assignments for document $d^*$ may be updated, but these are independent of all other documents' assignments in LDA. Therefore, the marginal value of all links that do not include $d^*$ are unchanged in the next iteration. These may be memoized \citep{iyer2019memoization} rather than recomputed, so each iteration we only need to recompute marginal values of at most $|\Phi|$ remaining links to $d^*$. Then, rather than check the maximum marginal valued link out of the $|\Phi||D|$ links to add to our solution each iteration, we save and update the best link for each document in a max-heap. Thus, we extract the best element in $O(\log |D|)$ instead of $O(|\Phi||D|)$.
    \item \textbf{Improving the per-iteration complexity by another $2|D|$-factor.} In a vanilla implementation, computing a \emph{single} marginal value of a new link $[\tau\Rightarrow d]$ requires summing over the log-probabilities of all $|D|$ documents. However, due to the conditional independence of documents in LDA, only the log probability of $d^*$ changes each iteration. After cancelling these terms from the difference $f_E([\tau\Rightarrow d]) = f(E\cup [\tau\Rightarrow d]) - f(E)$, we can compute the marginal value of $f_E([\tau\Rightarrow d])$ by summing only over the words in $d$, which accelerates computation by a factor of $2|D|$ each iteration.
    \item \textbf{Improving the per-iteration complexity by another $\kappa$-factor.} We can also memoize the log probabilities associated with the current topic-word assignments $z_{i,d}$. Then, when we compute a marginal value, we need not find the max log probability for each word among the $E_d$ topics connected to $d$, which saves an additional $\kappa$ factor per iteration.
    \item \textbf{Fast start.} Because documents in LDA are conditionally independent and each document must be drawn from at least one topic, we can efficiently initialize the set $E$ by adding a link between each document and the highest marginal value topic for that document. This yields the first $|D|$ links after computing a single round of marginal values (equivalent to one step of the for-loop instead of $|D|$ rounds). Moreover, this first round of marginal values can be efficiently computed as the product of the document-word matrix and $\Phi^T$ minus the phantom topic values (see the \textsc{FastInitialize} subroutine).
    \item \textbf{Lazy updates.} Our approach can also leverage lazy updates. Lazy updates offer no provable improvement in the worst-case, but often result in dramatic practical speedups \citep{minoux1978accelerated, mirzasoleiman2015lazier}. In particular, it is practically advantageous to lazily update the marginal values $M[d^*,:]$, as we do not need to know all marginal values, but just the best one, which in the best case removes an $O(|\Phi|)$ factor of computation each iteration. Also, when recomputing the marginal value of a link over various iterations, if the link does not improve the log probability of a word in an iteration then it will not improve the log probability of that word in any subsequent iteration, so we can skip the word in the marginal value summation.
    \item \textbf{Fast indexing via the heap.} In a practical fast implementation, each row of $\mathbf{P}$ and $\mathbf{M}$ can be stored in the heap $\mathbb{m}$ attached (e.g. in a tuple) to their corresponding elements to avoid the cost of indexing into $|D|$-sized storage.

\end{itemize}

\paragraph{\textbf{Complexity of \textsc{FastGreedy-E-LDA}}.}
To simplify analysis, we will regard the addition of each of the $|D|$ initial links added by \textsc{FastInitialize} as `iterations' of \textsc{FastGreedy-E-LDA}, despite the fact that these initial `iterations' are slightly faster. \textsc{FastGreedy-E-LDA} completes in $\kappa|D|$ iterations. Each iteration extracts the maximum-marginal-valued link $[t^*\Rightarrow d^*]$ from the max-heap $\mathbb{m}$, which is $O(\log |D|)$. The next line updates the current log probabilities corresponding to $d^*$'s current topic-word assignments $z_{d,1},\dots,z_{d,|d|}$, which is bounded by the wordcount of the longest document, $O(\ell)$. The total log probability of document $d$ is then updated in constant time (via memoization). Then the marginal values for all remaining links to $d^*$ are updated, and we also check which of these links has the greatest marginal value, which has complexity $O(\ell|\Phi|)$. Finally, we insert this link with the greatest marginal value for $d^*$ into the max-heap $\mathbb{m}$, which is $O(\log|D|)$. We have a total per-iteration complexity of just $O(\log |D| + \ell|\Phi|)$.

% In the subsequent appendix section, we show that \textsc{FastGreedy-E-LDA}'s runtime complexity improves on the fastest generic serial algorithm for submodular maximization, \textsc{Lazier-than-Lazy-Greedy}, despite the fact that \textsc{FastGreedy-E-LDA} obtains a deterministic near-optimal guarantee whereas \textsc{Lazier-than-Lazy-Greedy} obtains near optimal guarantees only in expectation.

\section{Comparison with Lazier-Than-Lazier-Greedy}
\label{sec:LTLGandcomplexity}
We now show that \textsc{FastGreedy-E-LDA} has better complexity than the fastest generic serial algorithm for submodular optimization, \textsc{Lazier-than-lazy-greedy} (LTLG) \citep{mirzasoleiman2015lazier}, despite the fact that \textsc{FastGreedy-E-LDA} obtains deterministically near-optimal solutions whereas \textsc{LTLG}'s guarantees hold only in expectation.

\begin{algorithm}[H]
%\caption{2-Class Directed Preferential Attachment}
\caption*{\textbf{Lazier-Than-Lazy-Greedy (LTLG)}}
\addcontentsline{loa}{algorithm}{LTLG}

\begin{algorithmic}
    	\INPUT  Corpus $D$, Candidate topic set $\Phi$,  Sparsity constraint upper-bound $\kappa$, Phantom topic links $\mathcal{P}$ \\
    	%\STATE \textbf{initialize} $\Phi_{MAP} \gets \{\phi_{phantom}\}$; topic-word assignments  $Z_{MAP}:z_{d,i} \gets \phi_{phantom}, \forall [d,i]$\\
     \STATE \textbf{initialize} $E \gets \mathcal{P}$ %topic-word assignments  $Z_{MAP}:z_{d,i} \gets \phi_{phantom}, \forall [d,i]$\\
     %\STATE \textbf{initialize} $E \gets \{[\tau_{phantom}\Rightarrow 1], [\tau_{phantom}\Rightarrow 2],\dots\, [\tau_{phantom}\Rightarrow |d|]\}$ %topic-word assignments  $Z_{MAP}:z_{d,i} \gets \phi_{phantom}, \forall [d,i]$\\
    	\STATE \textbf{for} $i \in \{1,\dots,\kappa|D|\}$ steps \textbf{do}\\
        \STATE \ \ \ \ $R \gets$ a random subset of remaining topic-document links $[\tau\Rightarrow d]$ not in $E$,  $\tau \in \Phi, d \in \D$\\
        % \STATE \ \ \ \ $\tau^* \gets$ the link $\tau$ in $R$ that adds the highest marginal value,  $\underset{\tau \in R}\argmax \  f_E(\tau)$\\
        \STATE \ \ \ \ \ $[\tau^*\Rightarrow d^*] \gets$ the topic-doc link in $R$ with the highest marginal value,  $\underset{\tau \in \Phi, d\in D}\argmax \  f_{E}([\tau\Rightarrow d])$\\
        %\STATE $Z \gets \argmax \ \big[ \sum_{d\in D}\sum_{i=1}^{|d|} \ \max_{\tau\in E_d} (\log \phi_\tau[w_{d,i}]) \big]$\\
        \STATE \ \ \ \ \ $E \gets [\tau^*\Rightarrow d^*]$
        % \STATE \textbf{return} \ \  $E \backslash \{[\tau_{phantom}\Rightarrow 1], [\tau_{phantom}\Rightarrow 2],\dots\, [\tau_{phantom}\Rightarrow |d|]\}$%,\ \  $Z$ 
        \STATE \textbf{return} \ \  $E \backslash \mathcal{P}$%,\ \  $Z$ 
        %
        %\STATE \textbf{return} \ \  $\Phi_{MAP} \backslash \phi_{phantom}$,\ $Z_{MAP}$
  \end{algorithmic}
  \label{alg:LTLG}
\end{algorithm}

\vspace{-2em}
\paragraph{\textbf{Complexity of \textsc{Lazier-Than-Lazy-Greedy} \citep{mirzasoleiman2015lazier}}.}
\textsc{LTLG} is the fastest generic probabilistic serial algorithm for maximizing a submodular function \citep{mirzasoleiman2015lazier}. It obtains a $(1-1/e - \epsilon)$ approximation guarantee \emph{in expectation}, where  $\epsilon$ is user-chosen. To accomplish this, LTLG proceeds identically to \textsc{Simple-Greedy}, except that at each of its $\kappa|D|$ iterations rather than evaluating the marginal value of all remaining links, it evaluates a random set of only $\frac{|\Phi||D|}{ \kappa|D|} log (1/\epsilon) = \frac{|\Phi|}{ \kappa} log (1/\epsilon)$ links sampled uniformly at random from the remaining links not yet added to the solution. It then adds the highest marginal value link in this random sample to the solution.

When applied to our objective $\dot{f}$, \textsc{LTLG} can also be accelerated via our memoization techniques ($\mathbf{P}$ and $\mathbf{p}$ in \textsc{FastGreedy-E-LDA}), which alleviate the need to sum over all $|D|$ documents' log probabilities and maximize over all already-linked topics for each marginal value. After applying these speedups, each marginal value can be computed at a cost of $O(\ell)$. Finding the best sample of all marginal values for an iteration is O($\frac{|\Phi|}{ \kappa} log (1/\epsilon)$) (i.e. the sample complexity per iteration). 

The result is a per-iteration complexity of $O((\ell|\Phi|/\kappa) \log (1/\epsilon))$ for each of the $\kappa|D|$ iterations. However, this ignores the additional complexity of drawing $\frac{|\Phi|}{ \kappa} \log (1/\epsilon)$ random samples each iteration, as well as the complexity of indexing into the memoized values each iteration. In practice, these random generation and indexing costs are significant, as the ground set is large. Also, unlike in the \textsc{FastGreedy-E-LDA} algorithm, the elements accessed each iteration by \textsc{LTLG} are non-proximal in memory (and associated with many different documents) due to the uniform random sampling, so there is no obvious way to apply the heap technique to improve indexing costs.

\bigskip \bigskip

\section{Deferred description of the \textsc{FAST} parallel algorithm}
\label{FAST_MODIFIED}
We reprint the \textsc{FAST-FULL} outer-loop and \textsc{FAST} algorithms from \citet{breuer2020fast} below using our E-LDA notation for completeness. As in that paper, the purpose of the outer loop algorithm is to generate various guesses $v\in V$ of the optimal constrained E-LDA objective value, which is denoted by $\OPT\coloneqq \max \dot{f}(E)$ s.t. $|E| \le \kappa|D|$. These guesses of $\OPT$ are then used to instantiate the inner \textsc{FAST} algorithm. This inner \textsc{FAST} algorithm then binary searches over these guesses for the largest guess $v$ that obtains a solution $E$ that is a $1-1/e$ approximation to $v$, resulting in an overall E-LDA solution that is near optimal.

%We show modifications to the original \textsc{FAST} algorithm in blue.

% \begin{algorithm}[H]
% \caption{\textsc{Fast-Full}: the full algorithm}
% \begin{algorithmic}
%     	\INPUT  function $f$, cardinality constraint $k$, parameter $\varepsilon$
%     	\STATE $V \leftarrow \geom(\max_{a} f(a), \max_{|S| \leq k} \sum_{a \in S} f(a), 1 - \varepsilon)$
%     	\STATE $v^{\star} \leftarrow$ \textsc{B-Search}$(\max\{v \in V: f(S_v) \geq (1 - 1/e)v\})$ 
%     	\STATE \qquad \ \  where $S_v \leftarrow \textsc{Fast}(v)$
%     	\STATE \textbf{return} $S_{v^{\star}}$ 
%   \end{algorithmic}
%   \label{alg:full}
% \end{algorithm}

\begin{algorithm}[H]
\caption{\textsc{Fast-Full}: the full algorithm}
\begin{algorithmic}
    	\INPUT  function $f$, cardinality constraint $\kappa$, accuracy parameter $\varepsilon$
     \STATE $S \leftarrow$ the highest value $\kappa|D|$ links $[\tau\Rightarrow d]$\\
    	\STATE $V \leftarrow \geom( f([\tau^* \Rightarrow d^*]), \max_{|S| \leq \kappa|D|} \sum_{[\tau \Rightarrow d] \in S} f([\tau \Rightarrow d]), 1 - \varepsilon)$
    	\STATE $v^{\star} \leftarrow$ \textsc{B-Search}$(\max\{v \in V: f(E_v) \geq (1 - 1/e)v\})$ 
    	\STATE \qquad \ \  where $E_v \leftarrow \textsc{Fast}(v)$
    	\STATE \textbf{return} $E_{v^{\star}}$ 
  \end{algorithmic}
  \label{alg:full}
\end{algorithm}

The main routine \textsc{Fast} \  generates at every iteration a uniformly random sequence $a_1, \ldots, a_{|X|}$ of the remaining set $X$ of topic-document links that have not yet been discarded. After the preprocessing step which adds to the current solution $E$ some topic-document links guaranteed to have high marginal contribution, the algorithm identifies the maximum sequence prefix length $i^*$ (i.e. the first $i^*$ links of the sequence) such that there is a large fraction of not-yet-added links in $X$ with high contribution to $E \cup A_{i^{\star} - 1}$, and adds this sequence prefix of topic-document links to the current solution $E$.

 \begin{algorithm}[H]
\caption{\textsc{Fast}: the Fast Adaptive Sequencing Technique algorithm}
\begin{algorithmic}
    	\INPUT  $f$, constraint $\kappa|D|$, guess $v$ for $\OPT$, accuracy parameter $\varepsilon$, sample complexity $m$
    	\STATE 	 $E \leftarrow \emptyset $
    	\STATE  \textbf{while} $|E| < \kappa|D|$ and number of  iterations  $< \varepsilon^{-1}$ \textbf{do}
    	\STATE   \ \ \ \ \  $X \leftarrow (\{[\tau \Rightarrow d]\}, \tau \in 1\dots|\Phi|, d\in D),  \hat{t} \leftarrow (1-\varepsilon) (v - f(E))/(\kappa|D|)$
	\STATE   \ \ \ \ \  \textbf{while} $X \neq \emptyset$ and $|E| < \kappa|D|$  \textbf{do}
    \STATE  \ \ \ \ \ \ \ \ \ \  \emph{(Denote single links $[t\Rightarrow d]$ by $\ a \ $ to simplify notation)}
		\STATE  \ \ \ \ \   \ \ \ \ \   $a_1, \ldots, a_{|X|} \leftarrow \sequence(X, |X|)$
		\STATE  \ \ \ \ \   \ \ \ \ \  $A_i \leftarrow a_1, \ldots, a_i$
	\STATE \ \ \ \ \   \ \ \ \ \   $E \leftarrow E \cup \{a_i : f_{E \cup A_{i-1}}(a_i) \geq \hat{t} \}$
	\STATE \ \ \ \ \   \ \ \ \ \ $X_0 \leftarrow \{a \in X : f_{E}(a) \geq \hat{t}\}$
	\STATE   \ \ \ \ \   \ \ \ \ \   \textbf{if} $|X_0| \leq (1 - \varepsilon) |X|$ \textbf{then} 
	\STATE   \ \ \ \ \   \ \ \ \ \  \ \ \ \ \   $X \leftarrow X_0$ and continue to  next iteration
	\STATE    \ \ \ \ \   \ \ \ \ \  $R \leftarrow \sample(X, {m})$, 
	\STATE    \ \ \ \ \   \ \ \ \ \ $I \leftarrow \geom(1, \kappa|D| - |E|, 1 - \varepsilon)$
%\STATE \ \ \ \ \  \ \ \ \ \  \ \ \ \ \   $I \leftarrow \geom(1, k - |S|, 1 - \varepsilon)$
\STATE  \ \ \ \ \  \ \ \ \ \ $R_i \leftarrow \left\{a \in R : f_{E \cup A_{i-1}}(a) \geq  \hat{t} \right\},$ for $i   \in I$
\STATE  \ \ \ \ \  \ \ \ \ \   $i^{\star} \leftarrow$ \textsc{B-Search}$(\max \{i  :  |R_i| \geq (1 - 2 \varepsilon) |R|\})$ 
	\STATE  \ \ \ \ \   \ \ \ \ \    $E \leftarrow E \cup A_{i^{\star} }$
    	\STATE \textbf{return} $E$
  \end{algorithmic}
  \label{alg:main}
\end{algorithm}

\vspace{-2em}
\section{Deferred analysis of simulating \textsc{Fast}}
\label{app:FASTalgorithmquerysimulation}

As described in Theorem \ref{theorem:FASTguarantees} \citep{breuer2020fast} and above, the parallel \textsc{Fast} algorithm obtains near-optimal solutions w.p. in a number of parallel rounds of objective function evaluations that is logarithmic in the number of documents. However, each E-LDA objective function evaluation in \textsc{Fast} is $O(\ell |\Phi||D|)$ in a vanilla implementation, which results in computationally infeasible runtimes for \textsc{Fast} in practice (note that, whereas \textsc{Simple-Greedy-E-LDA}  computes objective function evaluations on no more than $\kappa|D|$ links, \textsc{Fast} computes objective function evaluations on sets up to and including the entire ground set of $|\Phi||D|$ links). Our goal in this section is to describe how the same optimizations described for the serial algorithms above can also be applied to \textsc{Fast} when solving E-LDA. The result is a log-time parallel algorithm with near-optimal approximation guarantees where each parallel round of function evaluations is practically fast, with runtime that is independent of the size of the data $|D|$. Specifically, after applying speedups, each function evaluation has worst-case complexity bounded by either just $O(\ell)$ or $O(\ell|\Phi|)$.

 The key idea necessary to obtain fast simulated queries practically fast on the E-LDA objective is to observe that all steps of \textsc{Fast} except the sequence preprocessing and the binary search require function evaluations of the marginal values of \emph{single} topic-document links, not sets of links, to the current solution $E$. As such, for these steps of the \textsc{Fast} algorithm, the same memoization strategies described above may be applied.  Specifically, we memoize  (as in $\mathbf{P}_{|D|\times |V|}$ above) the log probability of each word in the data per the current iteration's topic-word assignments $Z$; we memoize (as in in $\mathbf{p}_{1 \times |D| }$ above) the current summed log probability of each document per the current $Z$; we memoize (as in $\mathbf{M}_{|D|\times|\Phi|}$ above) all current marginal values of links. Note that unlike in the serial \textsc{FastGreedy-E-LDA} algorithm, there is no need to memoize each document's best not-yet-added topic and its marginal value in a max-heap $\mathbb{m}$ (\textsc{Fast} does not seek the highest marginal valued link). By the same arguments as above, these memoizations permit the marginal value of any single link to be simulated in runtime just $O(\ell)$ as before.

 However, in sequence preprocessing steps ($E \leftarrow E \cup \{a_i : f_{E \cup A_{i-1}}(a_i) \geq \hat{t} \}$) and binary search ($i^* \leftarrow \textsc{B-Search} \dots$) steps, \textsc{Fast} evaluates the marginal value of singleton links to a union of the current solution $E$ and a new sequence $A$ of other links. Due to documents' independence in LDA, the marginal value of a link to document $d$ is independent of all links to other documents. A new sequence $A$ of links cannot contain more than $|\Phi|$ links to document $d$, so we have a worst-case complexity bounded by $O(\ell|\Phi|)$ to simulate a query of the marginal value of any link in these \textsc{Fast} steps. We note that because it is very unlikely that a randomly drawn sequence will contain all links to the same document, in expectation the runtime complexity of each simulated marginal value query is far less. We thus obtain a practically efficient fast parallel algorithm capable of solving E-LDA near-optimally for truly enormous information-age datasets.

\section{LDA for downstream causal inference: Topics as treatments}
\label{sec:causalbackground}

% We now show that unlike the LDA solutions returned by virtually all mainstream LDA solvers, E-LDA solutions can be computed while maintaining the independence and well-defined treatment assumptions necessary for downstream causal inference frameworks that seek to measure how texts’ topics cause outcomes, or how treatments cause texts that exhibit different topics. The result is a natural new method for making causal inferences using texts that inherits E-LDA's strong approximation and interpretability guarantees.

We review the main problems that prevent researchers from studying the topics inferred by mainstream LDA algorithms as treatments or outcomes in downstream causal inference frameworks:

\emph{1. Fundamental Problem of Causal Inference with Latent Variables (FPCILV).}
First, causal inference frameworks assume that the set of treatments is fixed and the treatment assignments are randomized. However, in LDA we infer the treatments (topics) from the dataset. This means that under a different random treatment assignment, the texts would be different, and thus the treatments (topics) we infer would also be different. As such, topics qua treatments are not well-defined. \citet{egami2022make} has recently described this problem as the Fundamental Problem of Causal Inference with Latent Variables (FPCILV).

\emph{2. Stable Unit Treatment Variable Assumption (SUTVA).}
Second,  causal inference frameworks such as Potential Outcomes \citep{rubin1974estimating} and Directed Acyclic Graphs (DAGs) \citep{pearl2009causality} require the data to conform to the Stable Unit Treatment Variable Assumption (SUTVA) \citep{rubin1980randomization}, %which is satisfied when the potential outcomes of an observation are independent of treatment assignment. 
which is satisfied when the response of an observation depends \emph{only} on its own treatment assignment, not other observations' treatment assignments. However, even when the documents themselves exhibit this independence, the topics that any mainstream LDA algorithm (Gibbs, Variational Inference, etc.) ascribes to one document depend on the other documents in the dataset, which violates SUTVA. Thus, documents' topic assignments that are inferred by mainstream LDA algorithms cannot be directly used by a downstream causal inference framework \citep{egami2022make}.

% \emph{3. Loss of statistical efficiency.}
% Third, there has been much excitement about techniques developed by \citet{egami2022make} to overcome FPCILV and SUTVA  problems. Specifically, \citet{egami2022make} attempt to circumvent these problems by (\emph{i}) randomly partitioning the documents in \emph{training} and \emph{test} sets, (\emph{ii}) running an LDA algorithm on the training set to infer the topics, then (\emph{iii}) applying a second round of different LDA inference algorithms that attempt to independently associate each test set document (\emph{out-of-sample}) with the topics learned from the training set. These techniques represent significant step forward for making LDA amenable to causal inference. Nonetheless, running causal inference only on test-set documents results in a significant loss of statistical efficiency. 

\emph{3. Out-of-sample LDA algorithms have unbounded mismeasurement, high computational cost, and lack interpretability.}
These techniques also highlight another problem:  To enforce FPCILV and SUTVA, the gold-standard approach splits documents into training/test sets, learns a topic model on the training set, then infers test set documents' topics using (separate) out-of-sample algorithms \cite{egami2022make}. However, standard LDA inference algorithms designed for out-of-sample inferences have unbounded mismeasurement in theory, and it is well known that they are biased and/or highly sensitive to random initializations in practice \citep{wallach2009evaluation, egami2022make}.\footnote{Specifically, mainstream routines to do out-of-sample inference for LDA include Harmonic Mean \citep{newton1994approximate}, Importance Sampling \citep{li2006pachinko} (incl. Empirical Likelihood \citep{mccallum2002mallet}), Chib-Style Estimation \citep{chib1995marginal}, Left-to-Right Evaluation \citep{wallach2008structured}, and Variational Inference-based methods \citep{blei2003latent, egami2022make}. One interesting exception with impressive theoretic properties is \citep{arora2016provable}, though its performance is known to be suboptimal on real data.} In other words,  even if the original LDA algorithm run on the training set correctly infers the topics and training set document mixtures, the (different) out-of-sample LDA algorithm run on test set documents may mismeasure documents' topics in the test set, leading to spurious causal inferences \citep{battaglia2024inference}. Mismeasurement is particularly problematic in this setting because most valid treatment effects in the social sciences are small. 

\emph{4. Topic mixtures are not well-defined treatments.}
Practitioners seeking to use documents' underlying topic mixtures $\theta_d$ as treatments have noted that such treatments are not well-defined for a separate reason \citep{fong2016discovery}. Specifically, documents' underlying topic mixtures must sum to $1$ (i.e. $||\theta_d||_1 = 1, \forall \theta_d$), so an observation (document) cannot have more [or less] of a topic-of-interest without a proportional decrease [or increase] in the various other topics, which may be accomplished in infinite ways.\footnote{In general, there is also another potential problem that prevents researchers from studying latent topics in a causal inference framework. Namely, there could exist unmeasured treatments (e.g. topics) that confound the measured treatment's effect \citep{fong2023causal}. However, we note it is standard to assume `no unmeasured confounders' in other causal inference settings, and LDA explicitly specifies the assumption conditional independences that obviate this problem. Moreover, the best known means to proceed absent this assumption has no provable guarantees in terms of identifying the correct topics, and it also lacks all of E-LDA's  interpretability and scalability properties \citep{fong2023causal}. We therefore argue that E-LDA offers an advantageous means to obtain causal inferences compared to this state-of-the-art approach.}

When topics are instead used as outcomes instead of treatments, analogous versions of many of these problems continue to apply \citep{modarressi2025causal}.

\section{MAP topic assignments are well-defined as treatments}
Rather than considering documents underlying topic mixtures $\theta_d$ as treatments (which are ill-defined as they must sum to 1), it is natural to consider topic-word assignments. For example, define the treatment as a document's count of words assigned to a topic of interest. This treatment is well-defined, as it is natural to imagine a counterfactual document that is shorter or longer (e.g. where fewer or additional words were added from a topic-of-interest), while holding constant the counts of words drawn from each other topic. Alternatively, one could instead define treatments as the presence or absence of a topic of interest in a document's MAP assignments.

\section{Exemplar-LDA for downstream causal inference}
\label{sec:causal}
We now describe how to modify our approach to perform out-of-sample inference such that we can infer the topic assignments for `test-set' (held-out) documents without violating FPCILV and SUTVA.  Specifically, we optimize the E-LDA objective subject to the modified sparsity constraint that no single document $d$ is generated by more than $|E_d|\le \kappa_d$ topics, where $\kappa_d$ is a user-chosen sparsity constraint that can be document-specific. Formally, we seek 
\begin{align}
\text{\emph{Out-of-sample objective $f^+$: } \ } &f^+(E_d) = \max \big[ \log P(Z_d|d,E_d) \big]  = \sum_{i=1}^{|d|} \ \max_{\tau\in E_d} (\log \phi_\tau[w_{d,i}])\notag\\%[-1em]
\text{find \ } \underset{{\{\tau\Rightarrow d\}, \tau \in \Phi}}{\argmax} &f^+(E_d) \text{\ subject to \ } |E_d|\le \kappa_d; \ \ \ \ |E_d|\ge 1
\label{eq:ETMOOS2}
\end{align}
In this case, it is natural to use information from the `training set' to inform the choice of the out-of-sample document's topic sparsity constraint $\kappa_d$. For example, this may be accomplished via a simple regression (e.g. Tobit regression) of the count of assignments in the training set solution documents as a function of their length. This predictor can then be applied to each held-out document to obtain an appropriate $\kappa_d$ for each test-set document.

\section{\textbf{Out-of-Sample-FastGreedy-E-LDA}}
\label{app:fastSUTVA}
For completeness, we now give the accelerated version of \textsc{Out-of-Sample-FastGreedy-E-LDA}, which applies the same speedups used in \textsc{FastGreedy-E-LDA} to out-of-sample inference of assignments for each document independently. The main difference is that \textsc{Out-of-Sample-FastGreedy-E-LDA} can be computed independently for each document (in parallel), and there is no need for a heap data structure to keep track of the best marginal value link across all documents. Note we explicitly write the loop over $|D|$ documents (which can be computed in parallel) as it is convenient to initialize all documents' first links at the same time.
%%%%%%%%%%%%%%%%%%%%%%   FastGreedy-E-LDA  %%%%%%%%%%%%%%%%%%%%%%
\begin{algorithm}[H]
\caption*{\textbf{Out-of-Sample-FastGreedy-E-LDA}}
\addcontentsline{loa}{algorithm}{Out-of-Sample-FastGreedy-E-LDA}

\begin{algorithmic}
    	\INPUT  Doc-word matrix $\mathbf{D}_{|D|\times|V|}$, topic matrix $\mathbf{\Phi}_{|\Phi| \times |V|}$, documents' sparsity upper-bounds $\{\kappa_d\}$ \\
     \STATE $E, \mathbf{P}, \mathbf{p}, \mathbf{M}, - \gets$ \textsc{ FastInitialize}$(\mathbf{D}, \mathbf{\Phi})$\\
        \STATE \textbf{for} each document $d$ \textbf{do} (in parallel)
         % \STATE \ \ \ \ \  $\mathbf{M_d} \gets$ \emph{heapify} the links in $\mathbf{M}$ that contain $\mathbf{d}$
         %
       %\STATE \ \ \ \ \  $\mathbb{m}_d \gets $ the topic in $\mathbb{m}$ for document $d$
        \STATE \ \ \ \ \ \textbf{for} $\kappa_d$ steps \textbf{do}
           %
           % Add the best topic-doc link to S
           \STATE \ \ \ \ \ \ \ \ \ \ $E_d \gets [\tau^*\Rightarrow d] \text{ where } t^* \text{ is the topic corresp. to } max \ \mathbf{M}[d,:]$\\
           %\STATE \ \ \ \ \ $E \gets [\tau^*\Rightarrow d]$\\
           %
           % Update P i.e. the curr_bestvals_perword_indoc_DxWidthIndexy. This is the values corresp. to Z.
           \STATE \ \ \ \ \ \ \ \ \ \ $\mathbf{P}[d,:] \gets \text{elementwise max of } \mathbf{P}[d,:]  \text{ and }\mathbf{\Phi}[t^*,:] \odot \mathbf{D}[d,:]   $
           %
           % update p i.e. curr_docvals
            \STATE \ \ \ \ \ \ \ \ \ \ $\mathbf{p}[d] \gets \mathbf{p}[d] + \mathbf{M}[d,t^*]$ \\%\textcolor{red}{\text{ Maybe remove this line and just do a prod with a $\mathbf{1}$ when we use \textbf{p} in \textbf{M} line below}}\\
           %
           % Update allmargvals_D_x_T
           \STATE \ \ \ \ \ \ \ \ \ \ $\mathbf{M}[d,:] \gets \mathbb{1}_{|V|\times 1} \cdot \text{ (elementwise max of }\mathbf{P}[d,:] \text{ and } \mathbf{\Phi} \odot [\mathbf{D}[d,:];\dots;\mathbf{D}[d,:] ]) - \mathbf{p}[d]$
           % its taking the Phi matrix and saying all rows (topics), the cols of the words in d*, maxed vs. the P for the best doc which is the current best,      summing those, and minusing out the curr docval of d*
           %
           % Update best_margval_topic_perdoc_idx,best_margval_topic_perdoc_value
           %\STATE \ \ \ \ \ \ \ \ \ \ $ \mathbb{m}_d \gets \max(\mathbf{M}_d) $
           %
           \STATE \ \ \ \ \ $E_d \gets  E_d \cup E_d $%,\ \  $Z$ 
           \STATE \textbf{return} \ $E $%,\ \  $Z$ 
\end{algorithmic}
\label{alg:SUTVAAccelGreedy}
\end{algorithm}
%$\Phi_R$ stands for realized; X is the random set;
%Where $\mathbb{1}$ is a column vector of $1$'s
%$\odot$ denotes the Hadamard (elementwise) product;

Importantly, these changes preserve the per-document near-optimal approximation guarantee that is relevant to causal inference applications (where it is advantageous to have a guarantee that each observation i.e. document is near-optimally measured). Finally, this approach conveys the additional advantage that each document can be optimized independently in parallel and we no longer pay the runtime cost of maintaining the max heap. This yields a parallel algorithm that not only preserves SUTVA independence, but also obtains this near-optimal solution in just $\max_d (\kappa_d)$ parallel iterations, each of improved complexity that is just $O( \ell|\Phi|)$.

\section{Deferred analysis of objective $\dot{f}$ with \textsc{Co-occurrence} topics}
\label{cooccurrence_proof}
We now show that with topics $\phi_{w^*}[v] \propto \exp(|D_{w^*,v}|+\epsilon)$ and $\theta[\tau] \propto \sum_{v\in V} \exp(|D_{w^*,v}|+\epsilon)$  where $\tau$ is the topic  generated about word $w^*,v$ then the E-LDA objective simplifies to an hierarchical extension of a familiar submodular objective function used in data summarization: $f(\mathtt{E}) = \max \big[ \log P(Z|D,\mathtt{E}) \big] = \sum_{d\in D}\sum_{i=1}^{|d|} \ \max_{\tau\in \mathtt{E}_d} (|D_{w^*,w_{d,i}}|)$.

\emph{Proof.}
\begin{align*}
    \text{P}(D, Z,\ | \alpha, \beta,  \Theta, \Phi ) &\propto 
        \prod_{d\in D} \prod^{|d|}_{i=1} Cat (w_{d,i} | z_{d,i})  \ Cat (z_{d, i} | \theta_d)  \\
         &=  \prod_{d\in D} \prod^{|d|}_{i=1} \Big(\frac{\exp(|D_{w^*,w_{d,i}}|+\epsilon)}{\sum_{v\in V} \exp(|D_{w^*,v}|+\epsilon)}\Big) \Big(\frac{\sum_{v\in V} \exp(|D_{w^*,v}|+\epsilon)}{ \sum_{\tau=1}^k \sum_{v\in V} \exp(|D_{w^*,v}|+\epsilon)}\Big)\\
         &\text{simplifying and taking logs:}\\
          \log \text{P}(D, Z,\ | \alpha, \beta,  \Theta, \Phi ) &\propto  \sum_{d\in D} \sum^{|d|}_{i=1}|D_{w^*,w_{d,i}}| +\epsilon - \log\Big( \sum_{\tau=1}^k \sum_{v\in V} \exp(|D_{w^*,v}|+\epsilon)\Big)\\
          &=  \sum_{d\in D} \sum^{|d|}_{i=1}|D_{w^*,w_{d,i}}| \ \ - const.
\end{align*}
\begin{align}
        \text{After dropping constants, the E-LDA MAP objective becomes:} \notag\\
        f(\mathtt{E}) = \max \big[ \log P(Z|D,\mathtt{E}) \big] = \sum_{d\in D}\sum_{i=1}^{|d|} \ \max_{\tau\in \mathtt{E}_d} (|D_{w^*,w_{d,i}}|)%\\\text{where $\tau$ is the topic associated with word $w^*$.}
\end{align}

\section{Experiments: Additional details}
\label{app:experiments}

\textbf{Datasets: Additional details.} We select the three datasets in these experiments because they reflect a diverse range of styles and genres that capture mainstream data of interest in social science applications. Reuters new article briefs \citep{lewis1997reuters} have a vocabulary of $|V|=24{,}035$ words and mean document length of $80.6$ words; US Congressional speeches \citep{thomas2006get} have a vocabulary of $|V|=13{,}857$ and a mean document length of $149.0$ (though document lengths vary widely in that dataset); and NewsGroups social media discussion board posts \citep{lang1995newsweeder} have a vocabulary of $|V|=20{,}254$  and a mean document length of $228.4$. 

For second set of experiments, we initialize $\Phi$ to contain one topic generated for each word  $w^*\in |V|$ for each of the topic generators.

\begin{figure}[t]
\centering
\begin{subfigure}[b]{0.9\textwidth}
\includegraphics[width=\textwidth]{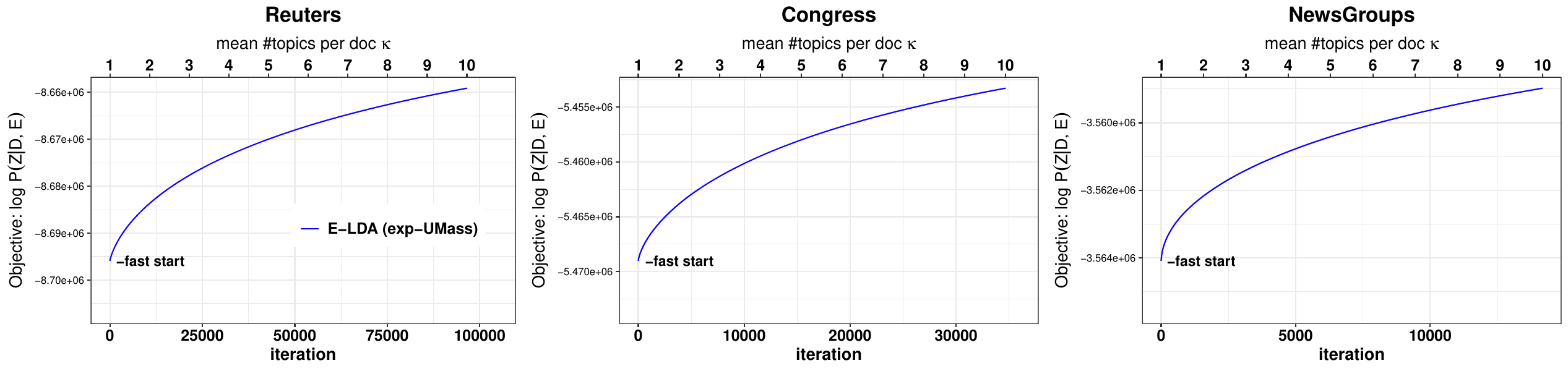}
   \caption[E-LDA convergence using \textsc{exp-UMass} topics]{Convergence of E-LDA's objective values using the \textsc{exp-UMass} topic generator.}
   \label{fig:convergence_expUmass}
\end{subfigure}
\begin{subfigure}[t]{0.9\textwidth}
\includegraphics[width=\textwidth]{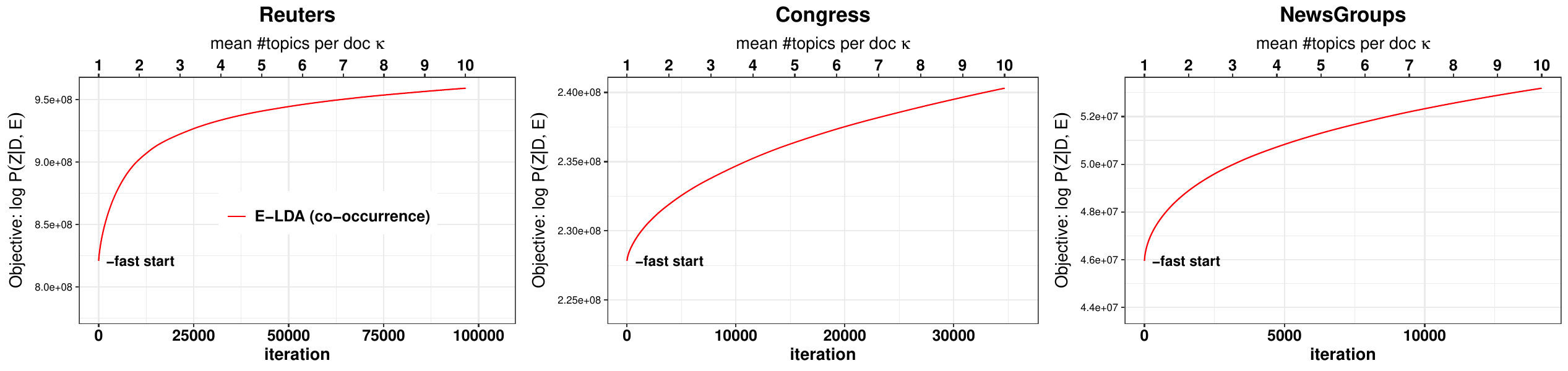}
   \caption[E-LDA convergence using \textsc{co-occurrence} topics]{Convergence using the \textsc{co-occurrence} topic generator.}
   \label{fig:convergence_coccurrence} 
\end{subfigure}
\caption[E-LDA convergence]{Convergence of E-LDA's objective values $\dot{f}$ using the (top row) \textsc{exp-UMass} topic generator and the (bottom row) \textsc{co-occurrence} topic generator.}
\label{fig:convergence}
\end{figure}

\textbf{Monotone convergence to near-optimal solutions.}
Figs. \ref{fig:convergence_expUmass} and \ref{fig:convergence_coccurrence} plot the convergence of E-LDA's objective values $\dot{f}$ using the (top row) \textsc{exp-UMass} topic generator and the (bottom row) \textsc{co-occurrence} topic generator. Note that unlike benchmarks, our algorithms converge monotonically as we increase the number of topics assigned to each document from a sparse $1$ per document (equivalent to a document \emph{clustering}) to a relatively dense $10$ per document (note the labels on top of the $x$-axes).

\begin{table}[h]
\centering
\tabcolsep=3.4pt
\begin{tabular}{@{}|l||lll|@{}}
    % \begin{tabular}{|r|p{2em}|p{3em}|p{3em}|r|p{2em}|p{2.5em}|p{2.5em}|}
\toprule
           & \textbf{Gibbs} & E-LDA & \color{gray}{rand$\kappa$}  \\ \midrule \midrule
Congress ($\times10^6$)   &   -2.05  (0.00834)  &    -1.78  (0.00631)    &     \color{gray}{-10.92 (0.0926)}       \\ \midrule
Reuters ($\times10^6$)   &   -2.80  (0.0114)  &   -2.52 (0.00948)   &   \color{gray}{-16.54 (0.226)}     \\ \midrule
NewsGroups ($\times10^6$) &     -1.57  (0.00536)  &    -1.40   (0.00450)   &     \color{gray}{-6.71 (0.0754)}    \\ \bottomrule
\midrule 
\toprule
           & \textbf{G-B} & E-LDA & \color{gray}{rand$\kappa$}  \\ \midrule \midrule
Congress ($\times10^6$)   &           -2.47  (0.00948)   &      -2.22   (0.00941)     &  \color{gray}{-10.22 (0.344)}      \\ \midrule
Reuters ($\times10^6$)   &       -3.66 (0.0107)     &   -3.43  (0.0128)  &  \color{gray}{-15.19 (0.126)}           \\ \midrule
NewsGroups ($\times10^6$) &       -1.70  (0.00399)   &      -1.61 (0.00428)       &  \color{gray}{-6.57 (0.0879)}        \\ \bottomrule
\midrule 
\toprule
           &  \textbf{G-S} & E-LDA & \color{gray}{rand$\kappa$} \\ \midrule \midrule
Congress ($\times10^6$)      &    -2.43    (0.00349)          & -2.14 (0.00338) &  \color{gray}{-10.31 (0.0156)}   \\ \midrule
Reuters ($\times10^6$)   &      -3.49   (0.0111)    &    -3.13  (0.00956) &   \color{gray}{-15.41 (0.170)}   \\ \midrule
NewsGroups ($\times10^6$)    &    -1.68    (0.00347)          & -1.55 (0.00247) &  \color{gray}{-6.52 (0.0770)}   \\ \bottomrule
\end{tabular}
\caption{Experiments set 1: Log posterior of assignments ($\dot{f}$ value) with same topics \& sparsity (10 run avg.), \emph{values reprinted from Table \ref{tab:OBJval} with added standard deviations in parentheses.} All means and standard deviations are scaled by $\times10^6$ for comparability.}
\label{tab:OBJvalSTDDEV}
\end{table}

\textbf{Experiments Set 1: Objective values over multiple experimental runs.}
Table \ref{tab:OBJvalSTDDEV} reprints the $\dot{f}$ values from Experiments Set 1 with standard deviations added in parentheses. We excluded these from the body of the paper in the interest of making each baseline's $\dot{f}$ values visually easier to compare across the table. For each baseline (Gibbs, Gibbs-Bayes, Gibbs-Sparse), \textsc{E-LDA}'s  objective values are higher for all $90$ experiments, and the difference for each benchmark and each dataset is statistically significant at the $p$$<10^{-10}$ level per standard $2$-sided t-tests.

\textbf{Experiments Set 2: evaluating the semantic quality of learned topics.} In Experiments Set 2, we are interested in comparing the semantic quality of topics generated by \textsc{E-LDA} to those generated by state-of-the-art Gibbs-based LDA solvers. We measure topic quality via the widely-used UMass coherence measure \citep{mimno2011optimizing, roder2015exploring}. Recall that UMass coherence measures the (log) probability that the top (i.e. highest-probability) $h^*$  words in a topic co-appear in documents in the corpus. Specifically, for a given topic $\tau$, the normalized coherence is:
\begin{align}
C(\tau,h^*)= \frac{2}{h^*(h^*-1)}\sum_{h=2}^{h^*} \sum_{\ell=1}^{h-1} \log \frac{|D_{\tau_h,\tau_\ell}| + \epsilon}{|D_{\tau_\ell|}}
\end{align}
where $D_{\tau_h}$ is the set of documents that contain the $h$'th most probable word according to topic $\tau$, $D_{\tau_h,\tau_\ell}$ is the set that contain both the $h$'th and $\ell$'th most probable words according to topic $\tau$, and $\epsilon$ is a small constant that precludes the $\log(0)$ case. We choose the normalized variant that includes the $\frac{2}{h^*(h^*-1)}\sum_{h=2}^{h^*}$ factor because it render coherence scores comparable across different choices of $h^*$, as in Figs. \ref{fig:manybenchmarks} ($x$-axes).

\begin{figure}
\centering
\includegraphics[width=\textwidth]{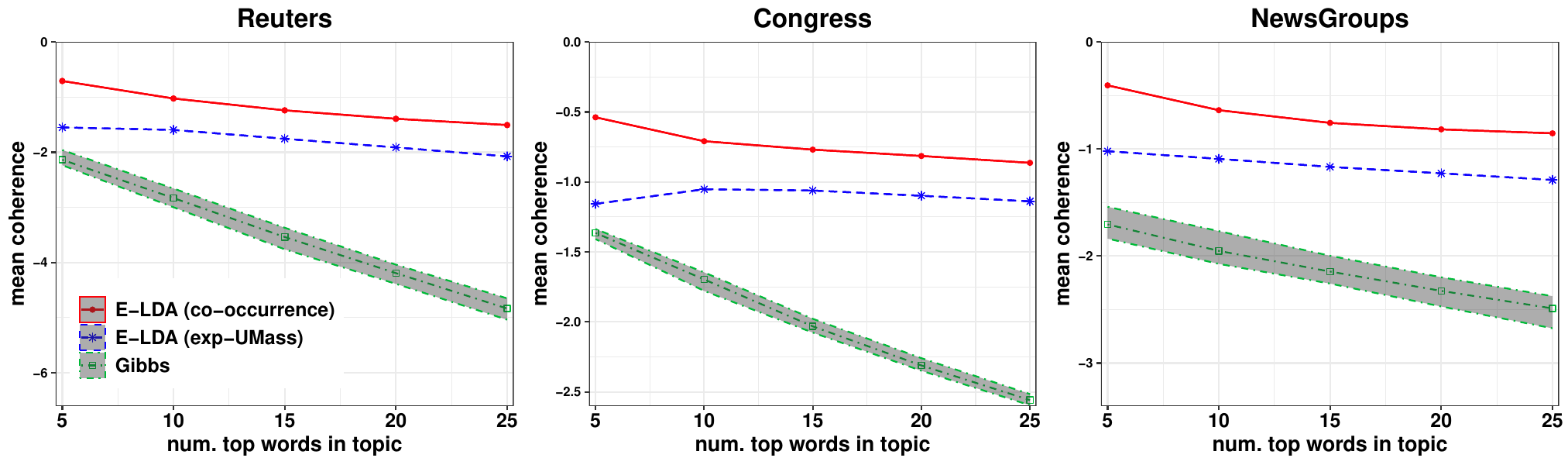}
   \caption{We reprint the main Gibbs baseline from Experiments Set 2 (Fig \ref{fig:manybenchmarks}) with added error bars showing its best and worst mean-coherence-per-run over the $10$ experimental runs.}
   \label{fig:coherence_errorbarGibbs} 
%\label{fig:convergence}
\end{figure}

\textbf{Experiments Set 2: Baseline coherence variation over multiple runs.} \textsc{FastGreedy-E-LDA} is deterministic, so its performance is always consistent across multiple runs. In contrast, Gibbs baselines' solution quality varies from one run to the next. Fig. \ref{fig:coherence_errorbarGibbs} reprints the main Gibbs baseline's mean coherence scores from Fig \ref{fig:manybenchmarks} with added error bars showing its best and worst mean-coherence-per-run over $10$ re-runs to show the range of its typical performance. Observe that even by re-running the baseline $10$ times, no single run obtains coherence comparable to either of the E-LDA solutions.

\textbf{Experiments Set 2: Additional baselines.}
For topic coherence experiments (Set $2$), we add recent neural and LLM-based baselines: \emph{BERTopic} \citep{grootendorst2022bertopic}, \emph{AVITM} \citep{srivastava2022autoencoding},  \emph{ETM}  \citep{dieng2020topic}, and \emph{FASTopic} \citep{wu2024fastopic}. We use the authors' codebases for each of these baselines (note that ETM and AVITM are now integrated with the \emph{OCTIS} library \citep{terragni2021octis}, and \emph{FASTopic} is now integrated with the \emph{Topmost} library \citep{wu2024towards}, and we set all parameters to be identical to the parameters in their respective papers. In terms of performance, \emph{ETM} is the best-performing baseline. This matches expectations, as \emph{ETM} was explicitly designed as a coherence-improving algorithm per \citep{dieng2020topic} (see also, e.g., \citet{wu2024survey}). We note that \textsc{AVITM}'s  poor coherence matches anecdotal experience that this algorithm  obtains good coherence on some datasets but requires customized parameter tuning to obtain competitive performance on others.

\textbf{Scale of experiments.} We note that the largest submodular problems in the literature are social network problems with ground sets of $\sim$$10^7$ elements. Our experiments described above consider $>10^8$ links. See \citet{breuer2020fast}, \citet{mirzasoleiman2013distributed}, and \citet{guestrin2005near}.

\section{Deferred discussion of unmodified \textsc{UMass} topic generator}
\label{app:umassbad}
We include the \textsc{UMass} topic generator both to build intuition, and also because its poor post-hoc coherence scores provide a clear illustration that other topic generators are not obtaining high post-hoc coherence merely due to their similarity with the UMass coherence function. In particular, we note that the unmodified \textsc{UMass} generator produces topics that each assign probability mass too uniformly across the vocabulary, resulting in un-sparse topics with unfocused semantic meaning. We note that a comparable property is referred to as "topic collapsing" when it arises in the context of recent neural topic modeling algorithms \citep{wu2023effective}.

To be clear, post-hoc coherence scores only weigh the ranking of words in a topic according to their probability mass, so order-preserving transformations of topic distributions have no effect on a topic's post-hoc coherence. Thus, the generated set of \emph{candidate} topics using the \textsc{UMass} generator has the same coherence score as \textsc{exp-UMass} and \textsc{Co-occurrence} topics. However, \textsc{UMass}'s too-uniform topics cause our algorithm to do a poor job of selecting (for each document) a small, diverse set of focused topics that each well-represent a subset of the document's words. This explains why the exponential transformations in \textsc{exp-UMass} and \textsc{Co-occurrence} produce much more focused topics, and also why the solution set of topics using these generators has far better post-hoc coherence scores than alternatives.

\section{\textsc{Co-occurrence} vs. \textsc{exp-UMass} topic generators in practice}
\label{app:cooccurvsumassinpractice}
We briefly note some practical experiences with the results of using \textsc{co-occurrence} vs. \textsc{exp-UMass} topic generators. Because \textsc{co-occurrence} favors topics associated with rigorous topic labels $w^*$ that are more popular, this results in fewer unique topics overall, and in topics that are more focused on familiar/popular concepts. In contrast, using \textsc{exp-UMass} topics tends to result in solutions that reflect more subtlety, including a wider variety of topics and less-common topic labels  $w^*$. In practice, we prefer to use \textsc{co-occurrence} as a `first cut' to familiarize ourselves with an unknown dataset, and then to apply \textsc{exp-UMass} to tease out subtler meanings for final analysis.

\section{Interpretability \& simulateability: Additional discussion}
\label{app:interpretabilityextra}
We briefly note a separate interpretability advantage of our approach that is particularly relevant to practitioners working on problem instances in social science and causal inference. Specifically, because all mainstream topic modeling algorithms are gradient-based, their routines are too complex for a human researcher to vet. Put differently, a practitioner cannot look into the algorithm's operations to obtain an answer to the question ``Why did the algorithm assign the `physics' topic to this document?". This aspect of interpretability is variously referred to as `simulateability' \citep{lipton2018mythos}. We note that this problem is also a focal point of recent social science methodology discussions---see, e.g., \citet{grimmer2022text}, \citet{ying2022topics}, and \citet{ke2024recent}.

In contrast, our Greedy-based algorithms are intuitively familiar to the human process of connecting topics to documents. Therefore, it is easy to answer such a question for a solution obtained by our algorithms. For example, the answer will be ``The `physics' topic was assigned to this document because it most improved the description of the document, which was previously described only by `space' and `particle' topics (i.e. the topics that were linked in $E_d$ before `physics')." By `most improved', we mean that the `physics' topic offered the greatest increase in the probability of that document's words compared to all remaining topics.

% \newpage

\end{document}